\documentclass{wfuthesis}

\usepackage[english]{babel}

\usepackage[letterpaper,top=2cm,bottom=2cm,left=3cm,right=3cm,marginparwidth=1.75cm,margin=1in]{geometry}

\usepackage{amsmath}
\usepackage{amsfonts}
\usepackage{graphicx}
\usepackage{subcaption}
\usepackage{booktabs} 
\usepackage{xcolor}

\usepackage{hyperref}
\hypersetup{
    colorlinks=true,
    linkcolor=blue,
    filecolor=blue,
    urlcolor=blue,
    citecolor=blue
}

\usepackage[margin=1in]{geometry} % set overall margins
\usepackage{multirow}
\usepackage{multicol}
\usepackage{hhline}
\usepackage{pdfpages}

\title{Fourier-enhanced Neural Networks For Systems Biology Applications}
\author{ENZE XU}
\date{May 2023}
\department{Computer Science}
\advisor{Minghan Chen, Ph.D.}
\chairperson{William Turkett, Ph.D.}
\member{Natalia Khuri, Ph.D.}

\begin{document}
\maketitle

\setstretch{2}
\acknowledgements{I would like to express my deep appreciation to all those who have supported me throughout my research project. Firstly, I would like to thank my parents for their unwavering love and encouragement. Their constant support has been a tremendous source of strength for me.

I would also like to extend my gratitude to my advisor, Dr. Minghan Chen, for her invaluable guidance and expertise. Her mentorship has been instrumental in shaping my research. I am also deeply grateful to Dr. William Turkett and Dr. Natalia Khuri for serving on my thesis committee and for their insightful feedback and suggestions.

Furthermore, I would like to thank my family, friends, significant other, and professors for their unwavering support and encouragement. I am also grateful to Mrs. Lesley Whitener for her invaluable administrative support within the Computer Science department. I would like to express my sincere appreciation to Mr. Cody Stevens for his assistance in managing the resources on the DEAC Cluster. I would also like to extend my thanks to all members of the Chen Lab group, whose collaboration has been invaluable to my research.

I am deeply grateful to Wake Forest University for providing the high-performance computing resources that have contributed significantly to my research results. Finally, I would like to express my gratitude to all faculty members of the Computer Science department, as well as all WFU staff, for their unwavering support throughout my academic journey.}

\setstretch{1}
\tableofcontents
\setstretch{2}

\listoffigures

\listoftables

\pchapter{List of Abbreviations}
\setstretch{1}
\begin{tabular}{ll}
AFNO & Adaptive Fourier Neural Operator \\
A-SIR & Age-structured SIR \\
BVP & Boundary Value Problem \\
CME & Chemical Master Equation \\
CNN & Convolutional Neural Network \\
CPINN & Conservative Physics-informed Neural Networks \\
DPINN & Distributed Physics-informed Neural Networks \\
ELU & Exponential Linear Unit \\
FDM & Finite Difference Method \\
FEM & Finite Element Method \\
FFT & Fast Fourier Transform \\
FNN & Fourier Neural Network \\
FNO & Fourier Neural Operator \\
FVM & Finite volume Method \\
GELU & Gaussian Error Linear Unit \\
GPINN & Gradient-enhanced PINN \\
IFNO & Implicit Fourier Neural Operator \\
IVP & Initial Value Problem \\
LAFNO & Linear Attention Coupled Fourier Neural Operator \\
LHS & Latin Hypercube sampling \\
MP-FNO & Model-parallel Fourier Neural Operator \\
mRNA & Messenger Ribonucleic Acid \\
MSE & Mean Square Error \\
NMSE & Normalized Mean Square Error \\
ODE & Ordinary Differential Equation \\
PDE & Partial Differential Equation \\
PFNO & Paralleled Fourier Neural Operator \\
PINN & Physics-informed Neural Network \\
ReLU & Rectified Linear Unit \\
SB-FNN & Systems-Biology Fourier-enhanced Neural Network \\
SciML & Scientific Machine Learning \\
Slurm & Simple Linux Utility for Resource Management \\
XPINN & Extended Physics-informed Neural Networks \\
\end{tabular}
\setstretch{2}

\begin{abstract}

In the field of systems biology, differential equations are commonly used to model biological systems, but solving them for large-scale and complex systems can be computationally expensive. Recently, the integration of machine learning and mathematical modeling has offered new opportunities for scientific discoveries in biology and health.

The emerging physics-informed neural network (PINN) has been proposed as a solution to this problem. However, PINN can be computationally expensive and unreliable for complex biological systems.

To address these issues, we propose the Fourier-enhanced Neural Networks for systems biology (SB-FNN). SB-FNN uses an embedded Fourier neural network with an adaptive activation function and a cyclic penalty function to optimize the prediction of biological dynamics, particularly for biological systems that exhibit oscillatory patterns. Experimental results demonstrate that SB-FNN achieves better performance and is more efficient than PINN for handling complex biological models.

Experimental results on cellular and population models demonstrate that SB-FNN outperforms PINN in both accuracy and efficiency, making it a promising alternative approach for handling complex biological models. The proposed method achieved better performance on six biological models and is expected to replace PINN as the most advanced method in systems biology.

\end{abstract}

\resetpagenum

\chapter{Introduction}
% \begin{equation}
% \mathcal{Z}^{\left(i\right)}=L^{\left(i\right)}\left(\mathcal{Z}^{\left(i-1\right)}\right)=\sigma\left( \mathcal{F}^{-1}_{M}\left(\mathcal{W}^{\left(i\right)}\times \mathcal{F}_{M}\left(\mathcal{Z}^{\left(i-1\right)}\right)\right) + F_{Res}^{\left(i\right)}\left(\sigma^{-1}\left(\mathcal{Z}^{\left(i-2\right)}\right)\right)\right),
% \end{equation}
% systems biology
Systems biology provides a framework for mathematically modeling and computationally simulating various biological systems across life science, which strives to interrogate complex biological processes and to understand the underlying molecular, biochemical, and physiological mechanisms that regulate cell behavior and population evolution.  
Generally, chemical reactions and molecular events occurring in a biological system can be systematically formulated into a set of ordinary or partial differential equations (ODEs, PDEs) as a function of time to characterize the temporal and spatial dynamics of genes, mRNAs, proteins, and cells at multiscale levels. 
Yet the promise of systems biology is impeded by its computational complexity, in which the intensive calculations and massive computing resources required to solve large-scale systems and perform quantitative analyses severely slow down the speed of model development and result interpretation \cite{bartocci2016computational, anantharaman2021stably}. 
It is even more challenging for spatial modeling that necessitates space discretizations, resulting in a trade-off between speed and accuracy: coarse grids are faster but less accurate, whereas fine grids are more precise but slower \cite{li2020multipole}. 
Therefore, this paper aims to develop a new computational tool that  enables computational biologists to apply model-based biological discovery to complex systems in an efficient and accurate manner.

This paper introduces a framework called the System-Biology Fourier-Enhanced Neural Network (SB-FNN) to solve the initial value problem in systems biology with high accuracy and efficiency. The motivation for this work includes two key factors: (i) the presence of ubiquitous and vital oscillatory patterns in biological systems, especially for molecular systems, which aligns well with the strong spectral periodicity characteristics underlying Fourier neural networks; (ii) the complexity of practical biological models, which often involve highly entangled variables and nonlinear functions that are challenging for conventional neural networks to handle accurately and efficiently.

%First, use the data to train the operator, and then use the physical equation to further optimize the operator. That is, it overcomes the shortcomings of pure data-driven (the data may be less or expensive to obtain) and also overcomes the shortcomings of pure PINN (low precision).

% Our main contribution is summarized as follows:

% \begin{itemize}
%     \item Our proposed SB-FNN incorporates two customized designs for systems biology applications: an embedded residual neural network (ResNet) and a cyclic penalty function that together optimize the prediction of dynamic behaviors of biological systems;
    
%     \item Compared with PINN, our proposed SB-FNN and its variants demonstrate a significant improvement in accuracy and efficiency, \chen{provide orders of magnitude} in both population and molecular dynamic models;
    
%     \item Compared with conventional neural networks, our SB-FNN has a simpler optimization environment and a more expressive representation space, which can solve practical complex biological models. 
    
% \end{itemize}

The main contribution is summarized as follows:
\begin{itemize}
    \item SB-FNN is a custom design for systems biology applications that builds upon the Fourier neural network (FNN) and incorporates two unique features: an adaptive activation function and a variance constraint. These two features work together to optimize the prediction of dynamic behaviors in biological systems.
    % \item Apply SB-FNN to both the initial value problem (IVP) and parameter estimation problem (PEP) in systems biology applications.
    \item Compared to PINN, our proposed SB-FNN and its variants exhibit significant improvements in accuracy and efficiency. SB-FNN provides a simpler optimization environment and a more expressive representation space, which enables it to solve practical and complex biological models.
\end{itemize}

\chapter{Background}
\section{Systems Biology}

System biology \cite{kitano2002systems,alon2019introduction} is an interdisciplinary field that aims to understand biological systems at the system level. It involves the integration of experimental and computational techniques to study the interactions and dynamics of biological components.

% \begin{figure}[h]
% \centering
% \includegraphics[width=1\textwidth]{fig/systems_biology.pdf}
% \caption[Systems biology circle] {Systems biology circle. The diagram illustrates the iterative process of the systems biology cycle, which involves both reality and mathematics steps. The image was sourced from open-access websites \cite{wikipediaepidemiology,covidcases}.}
% \label{fig:system_biology_circle}
% \end{figure}

% The systems biology circle \cite{palsson2015systems}, depicted in Figure \ref{fig:system_biology_circle}, represents the iterative approach used to model biological systems. 

The process of applying systems biology to a specific problem starts with identifying an issue within the biological system. This, in turn, leads to constructing a realistic model that captures the underlying structure of the system. This model is then mathematically modeled to obtain a solution for the initial problem. After obtaining a mathematical solution, it is validated by comparing it to experimental data. If necessary, the model is refined, and the cycle begins again. Moreover, interpreting the mathematical solution back into the real model provides insights and enhances the understanding of the system. Thus, the systems biology circle involves a continuous iteration between reality and mathematics, where each step builds on the previous one to gain a profound understanding of biological systems.

One crucial step in the systems biology circle is modeling, which includes creating mathematical models of biological systems. These models are essential in comprehending complex biological processes and predicting how biological systems behave under different conditions. Systems biology modeling has been applied to investigate multiple biological processes, including but not limited to biological circuits \cite{alon2019introduction}, bacterial networks \cite{covert2004integrating} and cell signal pathways \cite{kholodenko2006cell}.

Stochastic and deterministic modeling are two approaches used in system biology modeling \cite{srivastava2002stochastic}. The former considers the randomness in biological systems, while the latter assumes that the behavior of biological systems can be precisely predicted. Stochastic modeling is more appropriate when dealing with systems with small numbers of molecules, while deterministic modeling is suitable for larger systems.

Partial and ordinary differential equations are one approach in deterministic modeling used in system biology. Ordinary differential equations (ODEs) are used to model systems with continuous variables, such as time. In contrast, partial differential equations (PDEs) are used to model systems with more than one continuous variable, such as space and time. PDEs are particularly useful for studying systems that involve diffusion or transport, such as the spread of a signaling molecule in a tissue.

However, one of the main challenges in using differential equations in system biology modeling is the difficulty in solving them. Analytical solutions are often impossible to achieve, and numerical methods must be used to approximate the solutions. Additionally, the parameters and initial conditions used in the models must be estimated from experimental data, which can be challenging due to the complexity of biological systems.

\section{Scientific Machine Learning}

Scientific machine learning (SciML) \cite{rackauckas2020universal,roscher2020explainable} is a rapidly growing field that combines machine learning algorithms with the fundamental principles of science. In physics, SciML has the potential to revolutionize the way we understand and study complex physical systems. By leveraging the power of machine learning, researchers can extract valuable insights from large datasets, simulate complex physical processes, and even discover new phenomena that were previously unknown.

One of the most exciting applications of SciML in physics is the simulation of quantum systems \cite{rupp2015machine,torlai2020machine}. Quantum mechanics is notoriously difficult to understand and predict, but machine learning algorithms have shown promise in modeling and predicting the behavior of quantum systems. By training machine learning models on large datasets of quantum data, researchers can gain insights into the behavior of these systems and make predictions about their future behavior \cite{mehta2019high}. Another application of SciML in physics is in the field of cosmology \cite{villaescusa2021camels,carleo2019machine,mathuriya2018cosmoflow}. Cosmologists study the universe at the largest scales, and machine learning algorithms can help analyze the vast amounts of data collected by telescopes and observatories. By training machine learning models on large cosmological datasets, researchers can identify patterns and structures in the universe, process large 3D dark matter distribution \cite{mathuriya2018cosmoflow}, and even test fundamental physical theories like general relativity \cite{alestas2022machine}.

Scientific machine learning has the potential to transform the way we study and understand physical systems. As the field continues to develop and mature, we can expect to see even more exciting discoveries and breakthroughs in the years to come.

\section{Related Work}
%Traditional Approaches
The development of mathematical models and numerical methods for solving ordinary and partial differential equations has been a major focus of research for decades. These models and methods are widely used in many scientific fields, such as physics, engineering, and biology, to understand complex systems and make predictions about their behavior. Numerical approaches for solving differential equations, such as Euler's Method \cite{euler1952methodus}, Runge-Kutta Method \cite{runge1895numerische}, the finite element method (FEM) \cite{herrmann1967finite}, finite difference method (FDM) \cite{richardson1911ix} and finite volume method (FVM) \cite{eymard2000finite}, usually involve analytical or numerical techniques that rely on well-established mathematical theories.
These methods have proven to be effective and accurate for many problems, but they often require significant computational resources and can be limited in their ability to handle complex systems. In recent years, machine learning approaches have emerged as a promising alternative to traditional methods for solving differential equations. In this section, we will review both traditional and machine learning approaches for solving differential equations, with a focus on their applications in scientific problems.

% In this section, we will review machine learning approaches for solving differential equations, with a particular focus on physics-informed neural networks (PINN), which has demonstrated remarkable performance in various scientific fields and have emerged as a promising member of the machine learning family.

\subsection{Machine Learning Approaches}

Traditional numerical methods for solving differential equations often involve discretizing the space, which leads to a trade-off between computational efficiency and resolution. Finer grids offer better accuracy but can be computationally expensive, while coarser grids are faster but less accurate. To address these limitations, machine learning approaches such as Neural Ordinary Differential Equations (Neural ODEs) \cite{chen2018neural}, Deep Galerkin Method (DGM) \cite{sirignano2018dgm}, Physics-informed Neural Network (PINN) \cite{raissi2019physics}, and Fourier Neural Operator (FNO) have emerged as promising alternatives.

Neural ODEs represent dynamical systems using continuous-time ordinary differential equations, where the coefficients are neural network functions. They can learn complex dynamics from data even when the underlying system is not well understood. Neural ODEs have been successfully applied to a wide range of problems, including image processing \cite{chen2020mri} and time-series forecasting \cite{chen2022forecasting,gao2022explainable}.

DGM is a machine learning approach that uses deep neural networks to solve partial differential equations (PDEs). It approximates the PDE solution by replacing the finite-dimensional subspace in the Galerkin method with a neural network. DGM has shown promising results in solving various PDEs, including fluid dynamics \cite{li2022deep} and heat transfer problems \cite{zhang2022deep}.

In addition, Physics-informed Neural Network and Fourier Neural Operator are essential in the field of solving differential equations using machine learning. In the following subsections, we will discuss these methods in detail.

\subsection{Physics-informed Neural Network}
The Physics-informed Neural Network (PINN), a machine learning technique proposed for solving differential equations, has gained much popularity for its ability to achieve promising results for various problems in computational science and engineering~\cite{cai2022physics,misyris2020physics,cai2021physics,ji2021stiff,zanardi2022towards,lu2022deep,lv2021novel,sun2020surrogate,raissi2019deep,jin2021nsfnets,kissas2020machine,sahli2020physics}.

The primary objective of PINN is to utilize nonlinear partial differential equations to guide the supervised learning process. The model employs a deep learning network that can accurately identify nonlinear mappings from high-dimensional input and output data. The physical laws of the system act as a regulatory term, constraining the model to adhere to the governing equations. The authors successfully overcome the challenge of inadequate labeled data by using a large number of collocation points while obtaining favorable results in fluid dynamics, quantum mechanics, and reaction-diffusion problems. PINN is a meshless approach that transforms the initial and boundary value problems into optimization problems to solve differential equations.
That is, provided with explicit differential equations and initial and boundary conditions, PINN can solve a system without requiring labeled training data (ground truth solutions).
In the field of systems biology, two methods have been proposed as a domain translation of the PINN technique, such as 
biologically-informed neural network~\cite{lagergren2020biologically,greene2020biologically} and systems biology-informed deep learning~\cite{yazdani2020systems}.
Several real biological applications have taken advantage of PINN for model design and analysis, including soft biological tissue model~\cite{liu2020generic}, cardiac activation mapping~\cite{sahli2020physics}, and thrombus material properties~\cite{yin2021non}.

% While PINN has gained promising results across a wide range of problems in computational science and engineering \cite{sun2020surrogate,raissi2019deep,jin2021nsfnets,kissas2020machine,sahli2020physics}, it fails to achieve reliable training and correct solutions for complex systems and are prone to high computational costs \cite{krishnapriyan2021characterizing,lu2021learning}.

Although PINN has gained promising results across a wide range of problems in computational science and engineering \cite{sun2020surrogate,raissi2019deep,jin2021nsfnets,kissas2020machine,sahli2020physics}, there are two major deficiencies that hinder its broad application in biological and physical sciences: (i) it sometimes fails to achieve reliable training and correct solutions for complex systems~\cite{krishnapriyan2021characterizing, lu2021deepxde, meng2022physics}, and (ii) it is prone to high computational costs~\cite{anantharaman2021stably,jagtap2020locally,jagtap2020conservative,lu2021learning}.
To address these problems, researchers have been working on developing and optimizing the PINN algorithm.
A series of domain decomposition-based PINNs (CPINN, DPINN, XPINN) are presented to divide the computational domain into smaller subdomains~\cite{dwivedi2019distributed, jagtap2020conservative, jagtap2021extended}, thereby decomposing complex problems into smaller subproblems that can be solved by minimal size local PINN to solve.
The gradient-enhanced PINN (GPINN)~\cite{yu2022gradient} performs gradient enhancement on PINN by forcing the derivative of PDEs residual to be zero, which aids in faster convergence and improved accuracy of the model.
For small sampling data, few-shot learning is added to PINN, in which a neural network is used to train an approximate solution first and further optimized by minimizing the designed cost function~\cite{li2021deep}. 
% For linear problems, the Bayesian physical extreme machine learning is employed to enhance PINN, which increases the calculation speed by several orders of magnitude~\cite{liu2022bayesian}.
For stiff problems, Wang et al.~\cite{raissi2019physics} proposed a new network structure to deal with the numerical stiffness of the backpropagation gradient that causes PINN to fail.
Other methods focus on tailoring PINN architecture for a specific physical problem~\cite{wu2021modified, baddoo2021physics} and are difficult to generalize to systems biology applications.
% However, these techniques currently are difficult to  scale to the very large mo dels being used in practice by systems biologists. % copied
More research is needed to scale the generalizability and stability for these techniques to be applicable across the continuum of biological models.

% Fourier 
\subsection{Fourier Neural Operator}
Inspired by the Fourier neural networks (FNN) that apply the fast Fourier transform to the neural network~\cite{silvescu1999fourier}, a Fourier neural operator (FNO, a data-driven approach) is developed to learn a mapping of solutions to a complete family of differential equations~\cite{li2020fourier}. 
The Fourier transform reduces the computational complexity to quasi-linear, enabling large-scale calculations. 
Compared to traditional PDE solvers, FNO's resolution invariance property offers high accuracy at a low computational cost, making it a state-of-the-art approach for approximating PDEs.
Grady et al.\cite{grady2022towards} later break the limitation that FNO can only solve problems below three dimensions and mathematically proved that the proposed model-parallel Fourier neural operators (MP-FNO) can be used for dimensional data of any size. 
% And mathematically deduce all the necessary components of MP-FNO. MP-FNO is solving large-scale Parametric PDEs have a significant speed-up effect.
%Factorized Fourier Neural Operator (F-FNO)~\cite{tran2021factorized} has made a number of improvements on the basis of FNO to achieve better results. Factorize the input multi-dimensional Fourier transform and share the kernel integral operator parameters of all Fourier layers, which greatly reduces the amount of parameters. The residual structure added by F-FNO can greatly deepen the Fourier operator layers.
From the application perspective, FNO is customized to solve different problems, such as LAFNO~\cite{peng2022linear} for modeling 3D turbulence, PFNO~\cite{li2022solving} for solving seismic wavefield equations in velocity models, and IFNO~\cite{you2022learning} for predicting the mechanical response of materials.
In addition to learning challenging PDEs, FNO also achieved good results in the field of Vision transformers. The adaptive Fourier neural operator (AFNO)~\cite{guibas2021adaptive} draws on the idea of FNO and performs token mixing in the frequency domain through Fourier transform to understand the relationships among objects in a scene.
However, the above FNO-related methods are data-driven approaches that require ground truth solutions as labeled data to train the neural networks.

\chapter{Method}
\label{cha:method}

% \numberwithin{equation}{section}
% Structure

% Let $FFT$ denotes the Fast Fourier Transform of a function $f$ and $IFFT$ its inverse function, then

% $FFT_{f}\left(k\right)=\int _{D}\left( f_{j}\left(x\right)\cdot e^{-2i\pi\left<x,k\right>}\right) dx$,

% $IFFT_{f}\left(x\right)=\int _{D}\left( f_{j}\left(k\right)\cdot e^{2i\pi\left<x,k\right>}\right) dk$,

% Taking the xxxxx model as a guiding example ... 

\section{Fourier-enhanced Neural Network}
\label{cha:fnn}

To begin with, consider a general systems biology model, mathematically formulated as a PDE system:
\begin{equation}
y_{t}=\mathcal{G}\left[y\right](x), x\in \Omega,t\in \mathcal{T}
\end{equation}
\begin{equation}
y\left(\cdot, x\right) = g\left(x\right), x\in \partial\Omega
\end{equation}
where $\mathcal{G}$ is the differential operator for the PDE, $\Omega$ is the domain, $\mathcal{T}=\left[0,T\right]$ is the time domain, $\partial\Omega$ is the boundary of the domain, $x$ represents samples in the domain, $y\left(t,x\right)$ denotes the latent (hidden) solution, $g\left(\cdot\right)$ is a given boundary condition function.

\begin{figure}[h]
\centering
\includegraphics[width=1\textwidth]{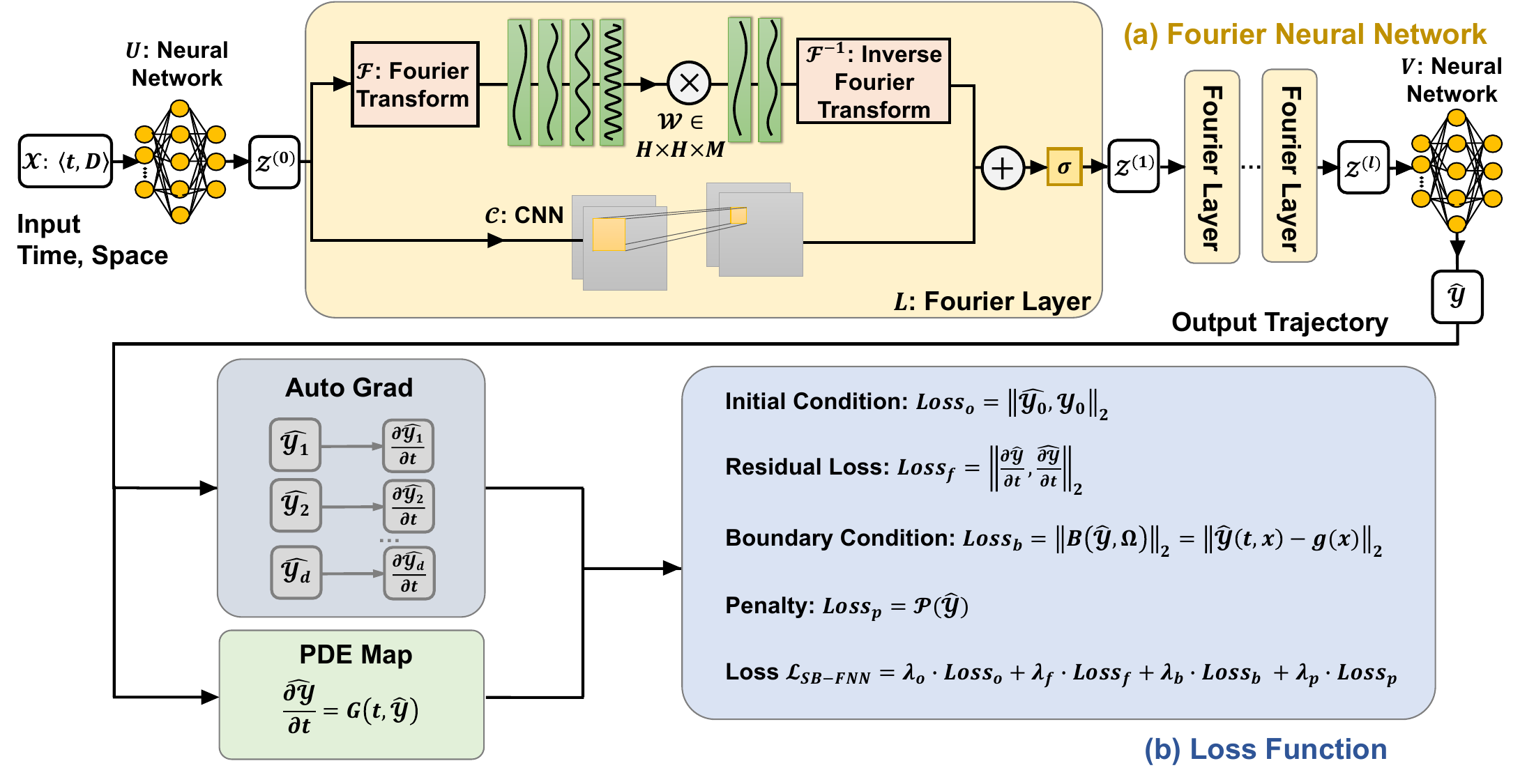}
\caption[Fourier-enhanced Neural Network for dynamic systems biology model prediction] {Fourier-enhanced Neural Network for dynamic systems biology model prediction. (a) The SB-FNN includes an input fully-connected neural network $\mathcal{U}$, $l$ Fourier layers, as well as an output fully connected neural network. (b) The loss function in the SB-FNN model is composed of four different loss components $Loss_{o}$ (initial condition), $Loss_{f}$ (residual loss), $Loss_{b}$ (boundary condition) and $Loss_{p}$ (physical penalty), each with its corresponding hyperparameter $\lambda_o$, $\lambda_f$, $\lambda_b$ and $\lambda_p$.}
\label{fig:SB-FNN}
\end{figure}

Let $\mathcal{X}\in \mathbb{R}^{T_{N}\times D}$ denotes the input of the Fourier neural network, where $T_{N}$ is the number of samples selected from time domain $\mathcal{T}=\left[0,T\right]$, and $D=\left|\Omega\right|$ is dimension of the spatial domain. The task of the network is to derive a stable estimation of $\mathcal{Y}\in \mathbb{R}^{T_{N}\times D}$, the stable latent solution. Motivated by ..., we propose FNN to map $\mathcal{X}$ to $\mathcal{Y}$:
\begin{equation}
\mathcal{Y}=SB\text{-}FNN\left(\mathcal{X}\right)=\left( V \circ L^{\left(l\right)} \circ \cdots \circ L^{\left(1\right)}\circ U \right)\left(\mathcal{X}\right),
\end{equation}
where $\circ$ is function composition, $ U$ is a fully-connected neural network that maps the input $\mathcal{X}$ into the initial hidden latent space $\mathcal{Z}^{\left(0\right)}\in \mathbb{R}^{T_{N}\times H}$, $H$ is the dimension of hidden latent space, $L^{\left(i\right)}$ indicates a Fourier Layer mapping the $\left(i-1\right)^{th}$ hidden latent space $\mathcal{Z}^{\left(i-1\right)}$ to the $i^{th}$ hidden latent space $\mathcal{Z}^{\left(i\right)}$, $ V$ is a fully-connected neural network that maps the final hidden latent space $\mathcal{Z}^{\left(l\right)}$ to the output $\mathcal{Y}$, and $l$ denotes the number of Fourier Layers.

Formulate each Fourier Layer $L^{\left(i\right)}$ as:
% Inspired by ..., we formulate each Fourier Layer $L^{\left(i\right)}$ as:
\begin{equation}
\mathcal{Z}^{\left(i\right)}=L^{\left(i\right)}\left(\mathcal{Z}^{\left(i-1\right)}\right)=\sigma\left( \mathcal{F}^{-1}_{M}\left(\mathcal{W}^{\left(i\right)}\times \mathcal{F}_{M}\left(\mathcal{Z}^{\left(i-1\right)}\right)\right) + C^{\left(i\right)}\left(\mathcal{Z}^{\left(i-1\right)}\right)\right),
\end{equation}

where $\sigma$ is a non-linear activation function, $C^{\left(i\right)}$ is a convolution neural network (CNN), $\mathcal{W}^{\left(i\right)}\in H\times H\times M$ is a weight matrix, $\mathcal{F}_{M}$ denotes the Fast Fourier Transform with $M$ as the number of Fourier modes to keep and $\mathcal{F}^{-1}_{M}$ is its inverse function. What we need to optimize are the parameters involved in the neural networks $U$, $V$ and the weight matrices $\mathcal{W}=\left\{\mathcal{W}^{\left(i\right)}\right\}^{l}_{i=1}$.

% $f\left(y\right)=u_{t}-\mathcal{G}\left[y\right](x)$, generally,
% \textbf{Solution to the initial value problem (IVP)}
To ensure the outputs of the neural networks follow the known biological mechanisms, the predicted $\widehat{y}$ is then brought into the corresponding multi-scale model as shown in Figure \ref{fig:SB-FNN}, represented as a set of partial differential equations with an initial condition $\widehat{y_0}=\widehat{y}|_{t=0}$ to calculate the distance from the model over a set of residual points over both time and spatial domain. This can be formulated as an optimization problem:
\begin{equation}
\min \limits_{\theta }\mathcal{L}_{SB-FNN} =
\lambda_{o}\left\| \widehat{y}_{0}-y_{0}\right\|_{2}
+ \lambda_{f}\left\| \widehat{y_{t}}-\mathcal{G}\left[\widehat{y}\right]\right\|
 +\lambda_{b}\left\| \widehat{y}\left(t,x\right)-g\left(x\right)\right\|_{2} 
 +\lambda_{p}\cdot \mathcal{P}\left(\widehat{y}\right),
\label{eq:loss}
\end{equation}

where $\theta$ denotes the parameters in neural networks $U$, $V$ and the weight matrices $\mathcal{W}$, $\mathcal{P}$ is an optimization penalty function used to restrict $y$ which will later be introduced, hyperparameters $\lambda_{f}$ controls the mismatch between the predicted gradient of $y$ and the gradient of $y$ derived by $\widehat{y}$ following the PDE equations, and $\lambda_{o}$, $\lambda_{b}$, $\lambda_{p}$ controls the discrepancy over boundary conditions, initial condition, optimization penalty, respectively. Figure \ref{fig:SB-FNN} (b) depicts the flow chart that demonstrates how the loss function is calculated.

\section{Adaptive Activation Function}
\label{cha:adaptive}

In deep learning, a neural network without an activation function can only be regarded as a linear model with a simple linear relationship. This type of model is relatively easy to solve, but it also lacks expressiveness and cannot solve complex problems. Therefore, activation functions are usually added to neural networks to increase their ability to express complex models by introducing nonlinear transformations. The types of activation functions are diverse, each with its own characteristics, which makes choosing the right activation function for a specific problem a crucial task in designing an effective neural network.

Tanh \cite{ackley1985learning}, short for hyperbolic tangent, is a popular activation function used in deep learning. It has an S-shaped curve and maps input values to output values ranging from -1 to 1. Mathematically, it can be expressed as
\begin{equation}
\label{eq:tanh}
\operatorname{Tanh}(x) = \frac{e^x - e^{-x}}{e^x + e^{-x}}.
\end{equation}
Tanh is zero-centered, which helps in optimization, and it is less prone to the vanishing gradient problem.

ReLU \cite{nair2010rectified} stands for rectified linear unit and is a widely used activation function in deep learning, which can be written as
\begin{equation}
\label{eq:relu}
\operatorname{ReLU} (x) = \max(0, x).
\end{equation}
ReLU is simple and computationally efficient, making it an attractive choice. It is a one-sided activation function, which means that it is activated only when the input is positive. ReLU solves the vanishing gradient problem that occurs in sigmoid by preventing the gradients from becoming too small, which leads to faster convergence. However, ReLU can face a problem called ``neuron death'' in which the neurons do not fire or respond to any input, leading to a dead end in the learning process.

Softplus \cite{glorot2011deep} is a smooth and differentiable activation function that is similar to ReLU. It maps input values to positive values, and its derivative is always positive. It can be written as

\begin{equation}
\label{eq:softplus}
\operatorname{Softplus} (x) = \ln(1 + e^x).
\end{equation}
Softplus is a popular alternative to ReLU because it does not suffer from the ``neuron death'' problem that occurs in ReLU. However, Softplus is not perfect and has its own disadvantages. It is not zero-centered, and its derivative at the origin is not defined, which can cause optimization problems.

The Exponential Linear Unit (ELU) \cite{clevert2015fast} is another variant of the ReLU activation function that has a smooth nonlinearity. ELU is defined as 

\begin{equation}
\label{eq:elu}
\operatorname{ELU} (x) =
\begin{cases}
x, & \text{if } x > 0 \\
\alpha (e^x - 1), & \text{otherwise}
\end{cases}
,
\end{equation}
where $\alpha$ is a hyperparameter that controls the slope of the function for negative inputs. ELU has been shown to outperform ReLU on many tasks and is less prone to the ``neuron death'' problem.

The Gaussian Error Linear Unit (GELU) \cite{hendrycks2016gaussian} activation function is a variant of the ReLU activation function that adds nonlinearity with a smooth curve. GELU is defined as 
\begin{equation}
\label{eq:gelu}
\operatorname{GELU} \left(x\right) = x\cdot P(X \leq x) = x\cdot \int_{-\infty}^{x} \frac{1}{\sqrt{2\pi}}e^{-\frac{t^2}{2}} dt,
\end{equation}

% \begin{equation}
% \label{eq:gelu}
% \operatorname{GELU} (x) = 0.5x(1+\operatorname{tanh}(\sqrt{2/\pi}(x+0.044715x^3)))
% \end{equation}

% \begin{equation}
% \Phi(x) = P(X \leq x) = \int_{-\infty}^{x} \frac{1}{\sqrt{2\pi}}e^{-\frac{t^2}{2}} dt
% \end{equation}
where $P$ is the cumulative distribution function of the standard normal distribution. The advantage of GELU is that it has both the sparsity-inducing property of ReLU and the smoothness of Sigmoid, making it more expressive and efficient than other activation functions.

The Sine (Sin) activation function is a periodic activation function that can be used to model periodic functions. The function is represented by the equation:
\begin{equation}
\label{eq:sin}
\operatorname{Sin^\ast} \left(x\right) = \operatorname{sin}\left(\beta \cdot x\right),
\end{equation}
where $x$ is the input to the function and $\beta$ is a scaling factor. The scaling factor $\beta$ is typically initialized to 1, but it is a trainable variable that can change during the training process.
The advantage of the Sin activation function is that it can capture the periodicity of the input and can be used to model complex periodic functions, such as those found in time-series data. However, the Sin activation function can be computationally expensive due to the trigonometric operations involved, and it is not suitable for all types of problems.

According to recent studies, incorporating periodic factors into activation functions can enhance the fitting ability of neural networks when predicting periodic functions \cite{sitzmann2020implicit}. This suggests that using a combination of multiple activation functions may be more effective in neural network design than relying on a single type of activation function. For example, GELU has been shown to outperform other activation functions such as ReLU and sigmoid in certain neural network architectures, and ELU has been shown to provide faster convergence and better accuracy than other activation functions in certain situations.

In neural network design, the choice of activation function and its hyperparameters can greatly affect the training effect of the model, and is typically solved by researchers based on experience and trial and error. This process is particularly important for PINNs, where the choice of activation function can be critical due to the different properties of the physical systems being solved. However, trying all activation functions in one round for each learning task is inefficient. A mixed activation function can be learned by selecting several general activation functions and performing linear weighted summation to obtain a mixed function. This approach can make the selection or even design of the activation function into an optimization problem, improving the efficiency and effectiveness of PINN modeling. By choosing the appropriate set of activation functions for each task and Fourier layer, better results can be achieved than using a fixed activation function for the entire model.

\begin{equation}
\sigma_{adaptive} = \sum_{j=1}^k \sigma_{j} * w_j
\end{equation}

\begin{equation}
w_i=Softmax(r)_i = \frac{e^{r_{i}}}{\sum_{j=1}^k e^{r_{j}}} \ \ \ 
\end{equation}

% \begin{figure}[h]
% \centering
% \includegraphics[width=0.5\textwidth]{fig/adaptive_1.pdf}
% \caption[Adaptive weights] {Adaptive weights. The adaptive weights, denoted as $w$, are generated by applying a Softmax operator to a trainable tensor vector, represented as $r$.}
% \label{fig:adaptive_weights}
% \end{figure}

\begin{figure}[h]
\centering
\includegraphics[width=1\textwidth]{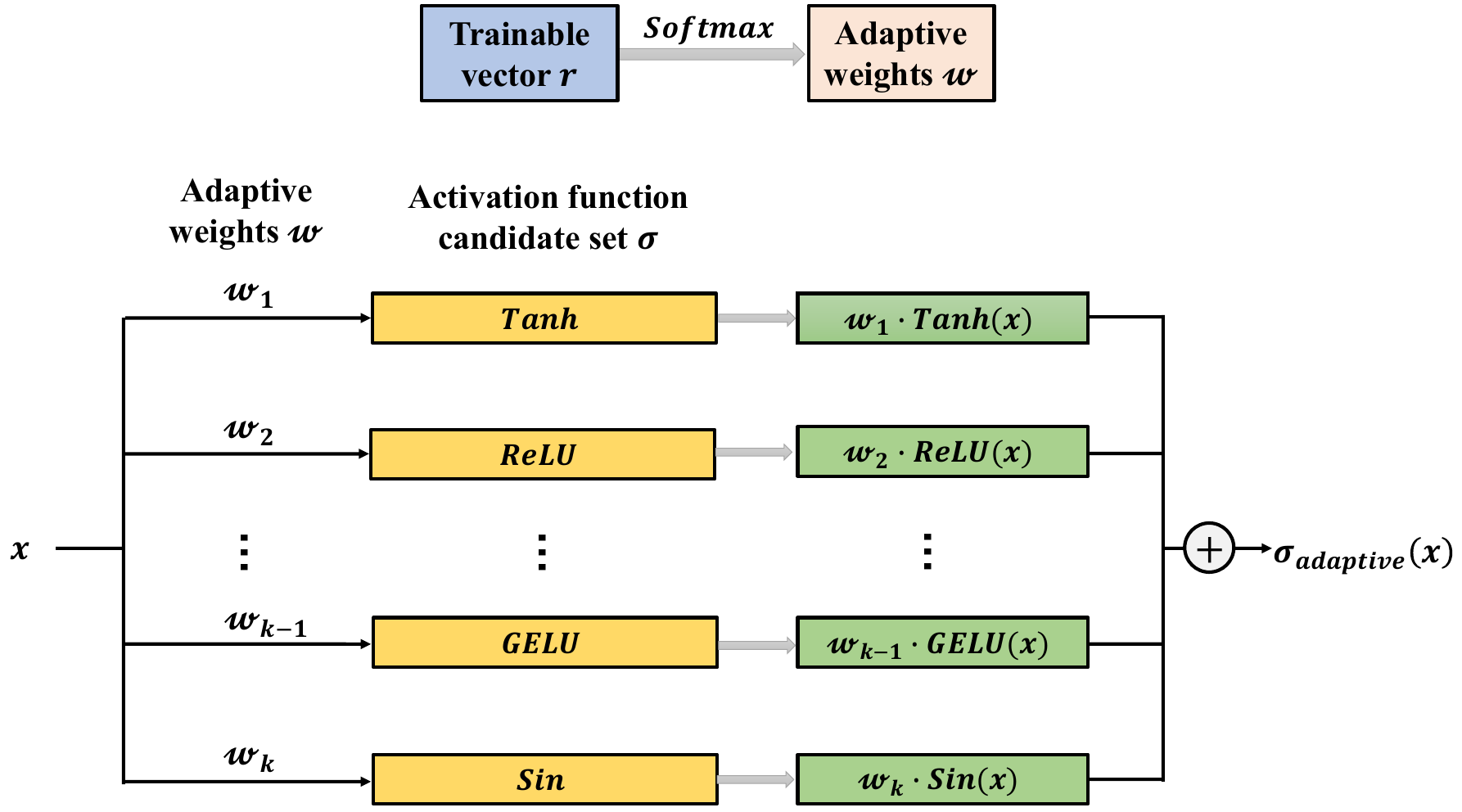}
\caption[Adaptive activation function] {Adaptive activation function. The adaptive weights, denoted as $w$, are generated by applying a Softmax operator to a trainable tensor vector, represented as $r$. In SB-FNN, Tanh, ReLU, Softplus, ELU, GELU, and Sin are chosen as the activation function candidates, as they have shown excellent performance across various models and tasks.}
\label{fig:adaptive_flow}
\end{figure}

Here $\sigma$ is the activation function candidate set, $k$ is the number of activation function candidates, $r$ is a trainable vector in length $k$ and $w$ is the transformed coefficient vector. In practice, we choose Tanh, ReLU, Softplus, ELU, GELU, and Sin as the activation function candidates, as they have shown excellent performance across various models and tasks. To obtain the optimal weighting coefficients for the activation functions, we employ a Softmax function to convert the trainable weight vector $z$ to the coefficient vector $w$. The flow of the designed adaptive activation function is shown in Figure \ref{fig:adaptive_flow}. By incorporating the mixed activation function into each Fourier layer, we replace the original GELU with a hybrid activation function, which allows each task and Fourier layer to have its own set of nonlinear transformations. This approach provides a more flexible and powerful way of nonlinear transformation, which can potentially improve the model performance.

We add a hybrid activation function to each Fourier layer with different trainable weight vector $z$, replacing the original GeLU, to optimize each hybrid activation function individually. The addition of a hybrid activation function to each Fourier layer offers a promising avenue for improving the performance and interpretability of deep learning models for physical systems. By leveraging prior knowledge about the system and carefully selecting the appropriate set of activation functions, we can build models that are both more effective and easier to understand.

\section{Variance Constraint}
\label{cha:cyclic}

\begin{figure}[h]
\centering
\includegraphics[width=0.7\textwidth]{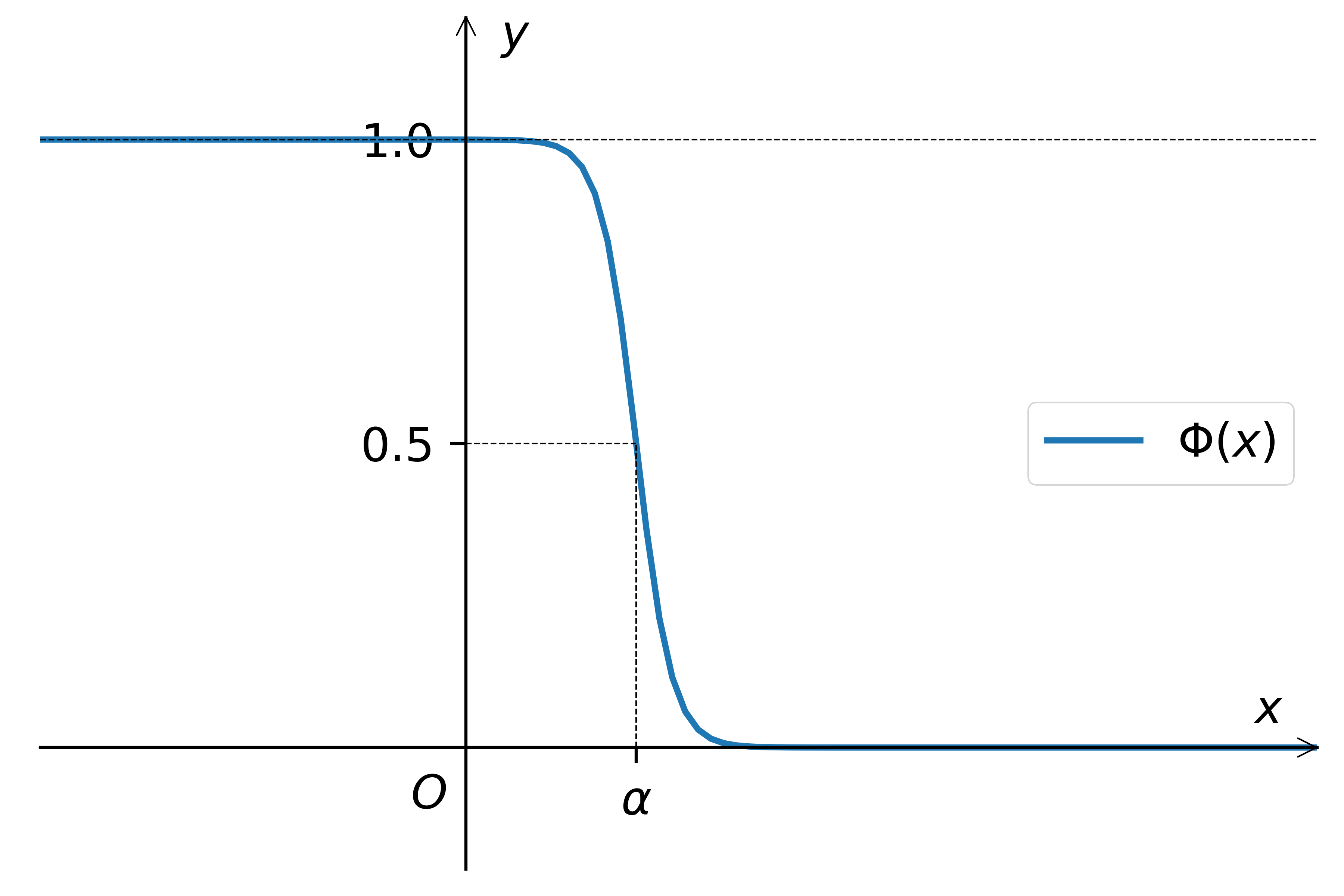}
\caption[Plot of the variance penalty function] {Plot of the variance penalty function $\Phi\left(x\right)$. The parameter $\alpha$ is utilized to manage the position of the symmetry center of the function along the x-axis, and the parameter $\tau$ is used to control the rate of change of the function around $x=\alpha$. A larger value of $\tau$ results in a faster decrease of the function from $y=1$ to $y=0$ in the neighborhood of $x=\alpha$.}
\label{fig:phi}
\end{figure}

Biological systems often exhibit ubiquitous and important oscillation patterns, especially for molecular systems. Therefore, we have developed a variance constraint, which is an oscillation penalty function, to help the model better obtain periodic characteristics:

\begin{equation}
Loss_{p}\left(\widehat{y}\right)=\lambda_{p}\cdot \mathcal{P}\left(\widehat{y}\right)=\lambda_{p}\cdot \sum^{D}_{i=1} \Phi\left(\operatorname{Var} \left( \operatorname{Norm} \left(\widehat{y}[i,:]\right)\right)\right)
\end{equation}

\begin{equation}
\Phi\left(x\right)=\dfrac{1}{2} \cdot (- \tanh\left(\left(x - \alpha\right) \cdot \tau\right) + 1)
\end{equation}
where $D$ denotes dimension of the spatial domain, $\lambda_{p}$ controls the weight of the variance penalty, $\operatorname{Var}\left(\cdot\right): \mathbb{R}^{n}\rightarrow \mathbb{R}$ is the variance of a vector, $\operatorname{Norm}\left(\cdot\right): \mathbb{R}^{n}\rightarrow \mathbb{R}^{n}$ is a normalization function that linearly normalizes a vector into a $[0,1]$ scale, $\Phi \left(\cdot\right): \mathbb{R}\rightarrow [0,1]$ represents a continuous oscillation-evaluation function used to punish low-variance curves, which involves a hyperbolic tangent function. When $x$ closes to $0$, it returns a value close to 1; While $x$ increases past a threshold $\alpha$, the function value reduces suddenly; When $x$ continues to increase, the function value gradually approaches $0$. The curve of the function $\Phi(x)$ in a plane is depicted in Figure \ref{fig:phi}. The hyper-parameters $\alpha$, $\tau$ need to be tailored to different systems biology models.

In summary, our designed variance constraint function provides an effective approach to capture the periodic nature of biological systems, especially for molecular systems. This function is continuous and utilizes a variance-based measure to penalize low-variance curves, which helps in improving the prediction accuracy of the neural network for oscillatory models. Additionally, this function is advantageous for the backpropagation process, as its derivatives are also continuous. We will later see from the experiment results how the incorporation of these factors into the model has resulted in significant improvements in prediction accuracy.

\chapter{Results}
% \section{Model-guided Fourier Neural Network}

This section showcases the effectiveness of our Systems-Biology Fourier-enhanced Neural Network (SB-FNN) in comparison to PINNs through six systems biology models. A comprehensive description of the systems biology models and their equations can be found in Appendix \ref{appendix:models}.

\section{Experiment Setup}

% Our experiments are performed on a multi-core Redhat Linux machine with 800GB RAM and two 32GB NVIDIA Tesla V100 GPUs. The dataset ($\mathcal{X}$ in Chapter 3) is randomly split into training (80\%) and testing (20\%) set. We set the hidden dimension ($H$ in Chapter 3) as 128, the number of Fourier modes as 32, the layer number as 4. 
% Those hyper-parameters can be revised according to the complexity and timescale of models.
% The normalized mean square error (N-MSE) is used to better evaluate the prediction accuracy across the time domain, written as

% Note that the N-MSE function is only used in evaluating how good the model can be used to predict the dynamic on the test dataset. Since the ground truth is not used in the training process, we use the loss function introduced in Chapter 3 instead.

We conducted our experiments on a multi-core Redhat Linux machine with two 32GB NVIDIA Tesla V100 GPUs. The hyper-parameters were set to a hidden dimension of 16, 12 Fourier modes, and a layer number of 4. As introduced in Chapter \ref{cha:method}, each of these FNN layers has a trainable independent weight vector to form the adaptive activation function with candidates ReLU, Tanh, Sin, GELU, ELU and Softplus. All these hyper-parameters can be adjusted based on the model's complexity and time constraints. Table \ref{tab:setup} displays the experiment setup for different models that was utilized in the following experiments.

% \begin{table}[htbp]
% \centering
% \caption{Experimental Setup}
% \begin{tabular}{l l}
% \hline
% \textbf{Parameter} & \textbf{Value} \\
% \hline
% Sample Size & 100 \\
% Treatment Group Size & 50 \\
% Control Group Size & 50 \\
% Treatment Method & Intervention X \\
% Control Method & Placebo Y \\
% Outcome Measure & Variable Z \\
% Data Collection Method & Survey \\
% Survey Instrument & Questionnaire \\
% Survey Respondents & Target population \\
% Statistical Analysis Method & ANOVA \\
% Significance Level & 0.05 \\
% \hline
% \end{tabular}%
% \label{tab:experiment_setup}%
% \end{table}

\begin{table*}[!htb]
\footnotesize
\caption[Experimental Configuration and Tuning] {Experimental Configuration and Tuning}
\label{tab:setup}
\begin{tabular}{ l cccccc } 
\toprule

Setup & Rep3 & Rep6 &  SIR & A-SIR & 1D Turing & 2D Turing \\
\midrule 
\midrule
Time domain & [0,10] & [0,20] & [0,100] & [0,100] & [0,10] & [0,2] \\
Maximum training epochs & 50,000 & 50,000 & 20,000 & 30,000 & 5,000 & 5,000 \\
Number of model parameters & 16,019 & 16,406 & 16,019 & 17,567 & 593,586 & 7,081,650 \\
Hidden layer size $H$ & 16 & 16 & 16 & 16 & 16 & 16 \\
Number of Fourier layers $l$ & 4 & 4 & 4 & 4 & 4 & 4 \\
Fourier modes $M$ & 12 & 12 & 12 & 12 & 12 & 12 \\
Initial learning rate & 0.01 & 0.01 & 0.01 & 0.01 & 0.01 & 0.01 \\
Optimizer & Adam & Adam & Adam & Adam & Adam & Adam \\
% Scheduler & \multicolumn{6}{c}{LambdaLR (Eq. \ref{eq:lr})} \\
Adaptive function candidates & \multicolumn{6}{c}{\{Tanh, ReLU, Softplus, ELU, GELU, Sin\}} \\
\bottomrule
\end{tabular}%}
\end{table*}

The dataset, denoted as $\mathcal{X}$ in Chapter \ref{cha:method}, was randomly sampled in the time domain following the Latin Hypercube sampling (LHS) method \cite{mckay1979two}. The LHS method is a type of stratified sampling that ensures that each point in the domain is sampled equally. It works by dividing the domain into equal-sized intervals and sampling a single point randomly from each interval. This ensures that each point in the domain is covered by exactly one sample and reduces the variance in the integral approximation. The dataset comprises a training set and a testing set, with a 10:1 length ratio. Both sets are sampled independently along the entire time domain, in ascending order, with the first time step always being zero. This ensures that the initial condition loss ($Loss_{o}=\lambda_{o}\left\| \widehat{y}_{0}-y_{0}\right\|_{2}$) and the PDE gradient loss ($Loss_{f}=\lambda_{f}\left\| \widehat{y_{t}}-\mathcal{G}\left[\widehat{y}\right]\right\|$) introduced in Eq. \ref{eq:loss} are properly functional.

In our study, we train our SB-FNN models using the Adam optimizer with a starting learning rate of 0.01. To further optimize our model's performance, we incorporated a learning rate scheduler
% , which was selected based on the provided configuration parameter. We offered support for four scheduler types, namely cosine annealing, step decay, decade decay, and fixed learning rate. To implement the cosine annealing scheduler, we utilized the CosineAnnealingLR scheduler with the maximum number of iterations set to the total number of training iterations. For the step decay scheduler, we applied the StepLR scheduler, with a step size of 5000 and a gamma value of $0.5$. 
following a decade decay strategy, which allows the learning rate (lr) to change with the number of epochs based on a scaling factor:
\begin{equation}
\label{eq:lr}
lr = lr\_init \cdot \dfrac{1}{\dfrac{ep}{b} + 1} = \dfrac{b\cdot lr\_init}{b+ep},
\end{equation}
where the parameter $lr\_init$ represents the initial learning rate, and $ep$ represents the number of epochs that has been run. The parameter $b$ is set to 1,000 for model Rep3, Rep6, SIR and A-SIR, and 100 for model 1D Turing and 2D Turing, as they have different maximum training epochs. As the epoch number increases, the learning rate decreases.

% Lastly, for the fixed learning rate scheduler, we set the learning rate to a constant value for all iterations. In our final result chart, we opted for the decade decay scheduler for all models due to its stability and fast convergence speed.

As the loss function depends on the prediction of the first time step ($y_0$ in Chapter \ref{cha:method}, the training process needs to use the entire prediction of $y$ in each epoch to calculate the loss. This means that, following the PINN-like design of the loss function, each epoch will involve one iteration with the complete training dataset as the batch size.

To guarantee convergence of each model, we have set a uniform number of epochs to 50,000 despite the fact that some of the models may converge earlier.

To assess prediction accuracy across the time domain on the test dataset, we used the normalized mean square error (N-MSE), which can be written as:

\begin{equation}
\label{eq:nmse}
\text{N-MSE}=\dfrac{1}{n}\sum^{n}_{i=1}\dfrac{\left\| \hat{\psi_{i}} - \psi_{i}\right\| _{2}}{\left\| \psi_{i}\right\| _{2}},
\end{equation}
where $\left\| \cdot\right\| _{2}$ is the 2-norm, $n$ is the batch size, and $\hat{\psi}$ is the prediction of the ground truth $\psi$. It is important to note that the N-MSE function was only used to evaluate the model's predictive performance on the test dataset, as the ground truth was not utilized during the training process. Instead, we use a loss function introduced in Chapter \ref{cha:method} to measure the model's performance during training. Additionally, for models such as SIR and A-SIR, which have dynamics that eventually reach a stable state of 0, this can cause the N-MSE equation's denominator to approach 0, resulting in unexpected results. To avoid this issue, we exclude these dynamics from the calculation of the N-MSE loss and only consider dynamics that do not reach 0 as the stable state.

\section{Prediction of Model Dynamics}

Systems biology systems are a class of mathematical models that have found applications in many areas, including biology, physics, and chemistry. They describe the dynamics of complex biological processes such as gene expression, cell signaling, and metabolic pathways. Many of these systems exhibit interesting phenomena, such as oscillations, stable states, and stiff rising and reducing behaviors.

In this paper, we investigate three common categories of systems: the Repressilator model, SIR model, and Turing Patterns. For each category, we examine two variations: the Repressilator model with only protein equations or with both mRNA and protein equations, the SIR model and its age-structured variant, and the Turing Pattern on both 1D and 2D spaces. A comprehensive explanation of each model is provided in Appendix \ref{appendix:models}.

% We have selected three typical classes of systems from a large number of systems biology PDE systems: the Repressilator model, SIR model, and Turing Patterns. For each of these classes, we consider two variants, namely the Repressilator model with protein equations only or both protein and mRNA equations, the SIR model and its age-structured variant, and the Turing Pattern on both 1D and 2D grids. The detailed description of each model is available in Appendix \ref{appendix:models}.

By training each model for a sufficient number of epochs using the proposed SB-FNN, accurate predictions of the dynamics of each model on the training set can be achieved.
% The corresponding model parameters for each curve with their respective accurate predictions are presented in Figure \ref{fig:fine_prediction}.

\subsection{Repressilator: Protein Only Model}
\begin{figure}[h]
\centering
\includegraphics[width=1.0\textwidth]{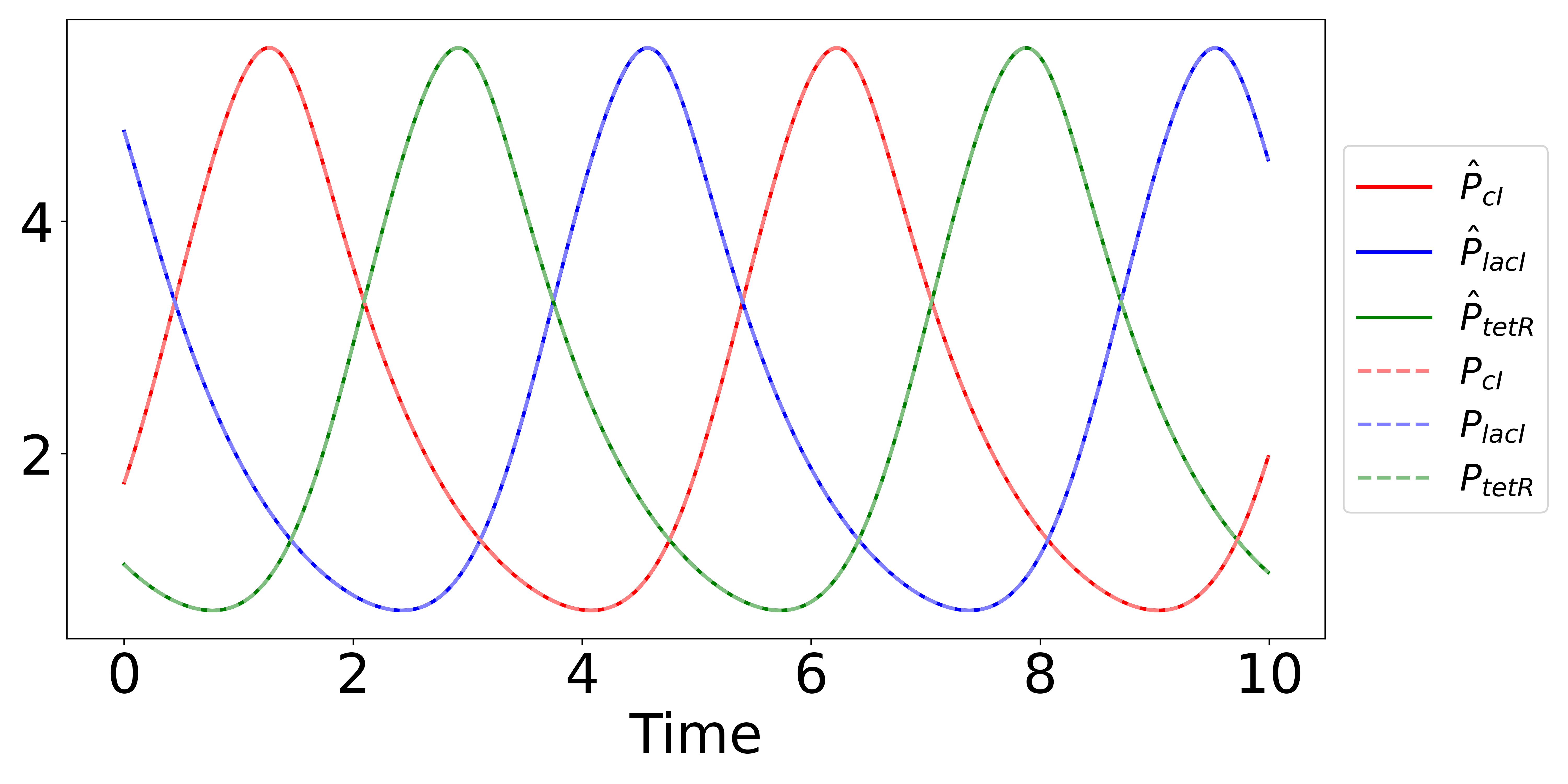}
\caption[Prediction of the Repressilator: protein only model] {Prediction of the Repressilator: protein only model. $P_{i}$ ($i=$ $lacI$, $tetR$, or $cI$) represents the concentrations of the repressor-proteins of $lacI$, $tetR$, and $cI$, respectively. The model parameters are $\beta = 10$ and $n=3$.}
\label{fig:rep3_prediction}
\end{figure}

The repressilator model \cite{elowitz2000synthetic} describes a synthetic oscillatory system of transcriptional repressors. Appendix \ref{appendix:rep3} contains the complete definition and equations of this model. If we only consider the protein interactions, there would be three variables in this model (Rep3), which represent the concentrations of the repressor-proteins of $lacI$, $tetR$, and $cI$. The prediction results of this model using SB-FNN are shown in Figure \ref{fig:rep3_prediction}. The variables with hats represent the predicted values, while those without hats represent the ground truth values, which are calculated. This definition applies to subsequent sections as well.

\subsection{Repressilator: mRNA and Protein Model}
\begin{figure}[h]
\centering
\includegraphics[width=1.0\textwidth]{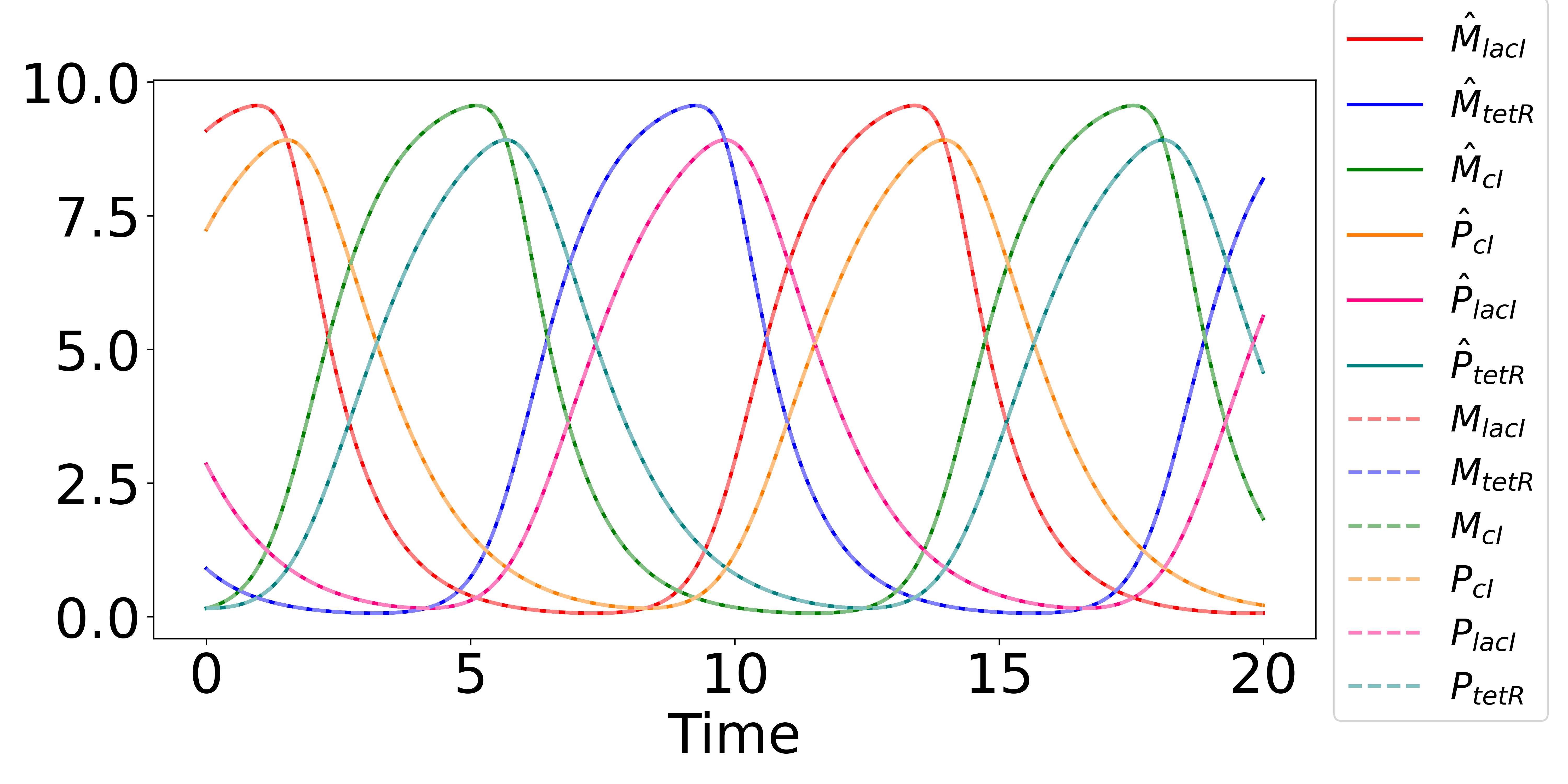}
\caption[Prediction of the Repressilator: mRNA and protein model] {Prediction of the Repressilator: mRNA and protein model. $P_{i}$ and $M_{i}$ ($i=$ $lacI$, $tetR$, or $cI$) represents the repressor-proteins concentrations and mRNA concentrations of $lacI$, $tetR$, and $cI$, respectively. The model parameters are $\beta = 10$, $n=3$, $\alpha=10$ and $\alpha_0=1e^{-5}$.}
\label{fig:rep6_prediction}
\end{figure}

Taking both transcription and translation reactions among three pairs of repressor mRNAs and proteins, the Repressilator model can be extended to differential equations with six variables (Rep6). Appendix \ref{appendix:rep6} contains the complete definition and equations of this model. Here $P_{i}$ denotes the repressor-protein concentrations, and $M_{i}$ represents the corresponding mRNA concentrations, where $i$ is $lacI$, $tetR$, or $cI$. The prediction results of this model using SB-FNN are shown in Figure \ref{fig:rep6_prediction}.

\subsection{SIR Model}
\begin{figure}[h]
\centering
\includegraphics[width=1.0\textwidth]{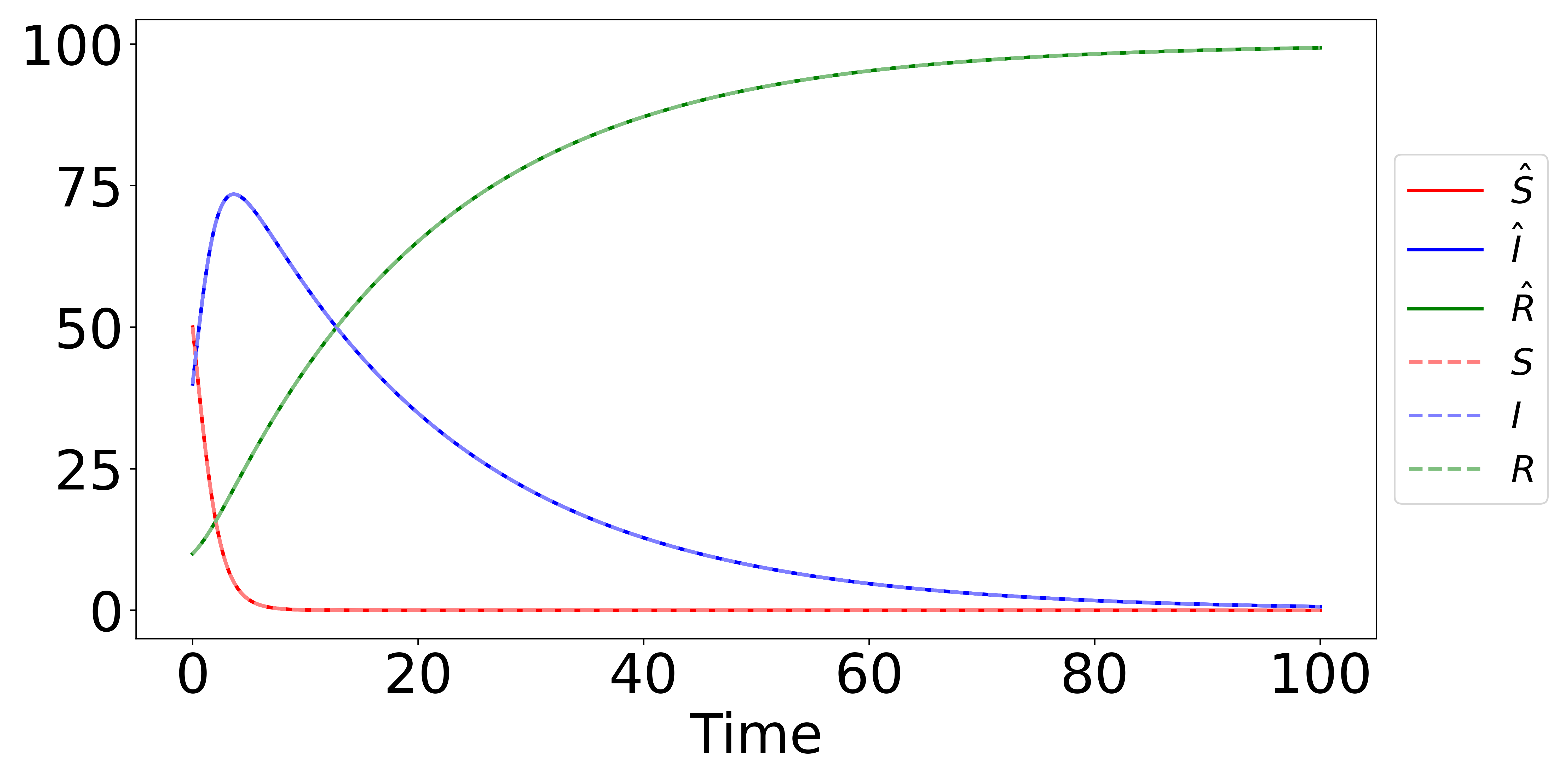}
\caption[Prediction of the SIR model] {Prediction of the SIR model. $S,I,R$ represents the susceptible, infectious, and recovered population, respectively. The model parameters are $\beta = 0.01$, $\gamma=0.05$ and $N=100$.}
\label{fig:sir_prediction}
\end{figure}

The SIR \cite{anderson1991discussion} is a classic compartmental model in epidemiology to simulate the spread of infectious disease. The basic SIR model considers a closed population with three different labels susceptible (S), infectious (I), and recovered (R). Appendix \ref{appendix:sir} includes the complete definition and equations of this model. The prediction results of this model using SB-FNN are shown in Figure \ref{fig:sir_prediction}.

\subsection{Age-structured SIR Model}
\begin{figure}[h]
\centering
\includegraphics[width=1.0\textwidth]{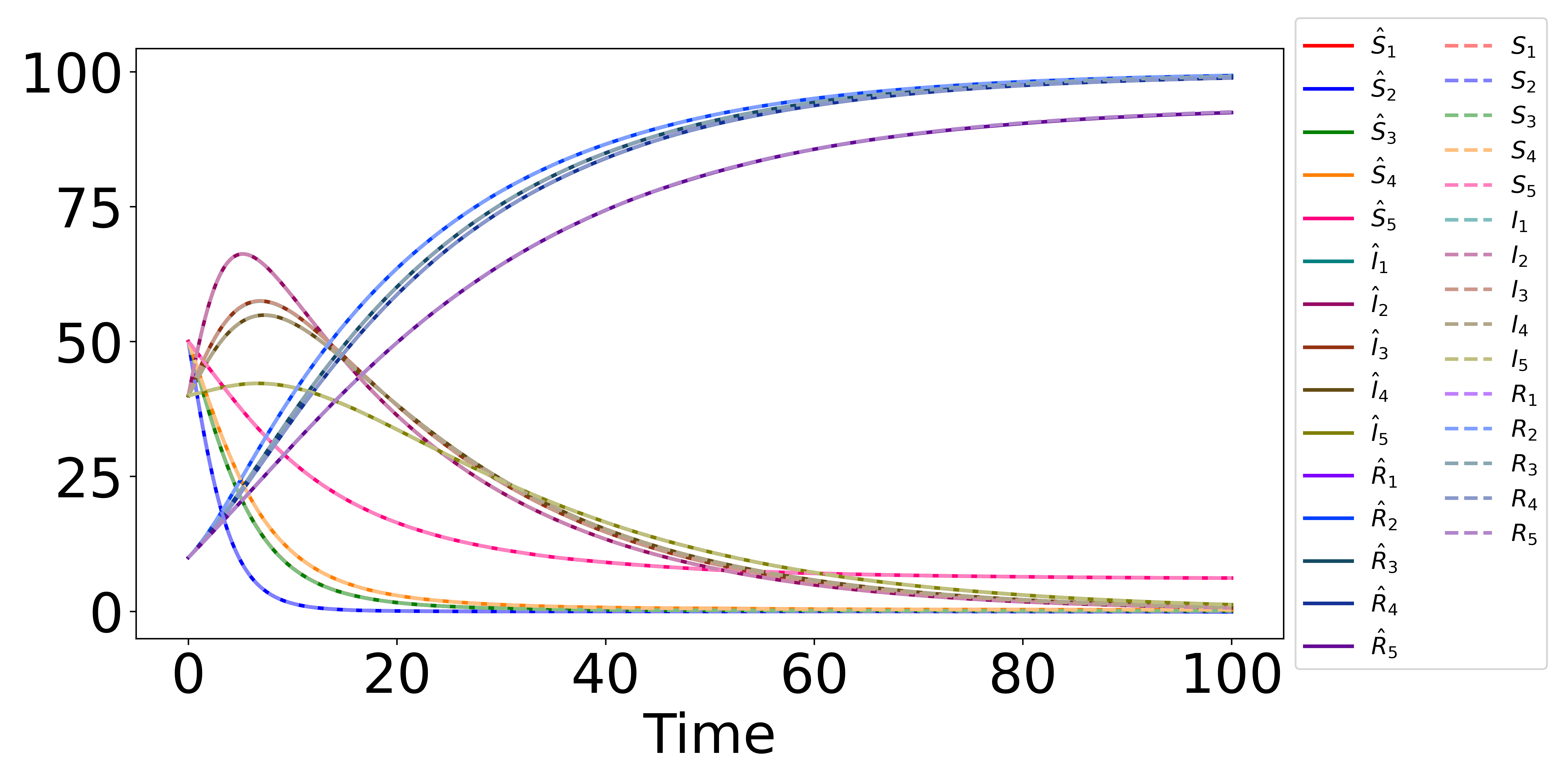}
\caption[Prediction of the Age-structured SIR model] {Prediction of the Age-structured SIR model. $S_i$, $I_i$, and $R_i$ represent the susceptible, infectious, and removed population within the $i$th age group, respectively. The model parameters are $\beta = 0.01$, $\gamma=0.05$, $N=100$, $n=5$, and $\mathcal{M}$ is an age-content matrix from \cite{ram2021modified}.}
\label{fig:asir_prediction}
\end{figure}

The Age-structured SIR (A-SIR) model, which takes into account differences in age groups \cite{ram2021modified}, is a variation of the SIR model. The variables $S_i$, $I_i$, and $R_i$ represent the susceptible, infectious, and removed population within the $i$th age group, where $1\leq i \leq n$, and $n$ represents the total number of age groups. Appendix \ref{appendix:asir} contains the complete definition and equations for this model. The prediction results for this model using SB-FNN can be seen in Figure \ref{fig:asir_prediction}.

\subsection{1D Turing Model}
\begin{figure}[h]
\centering
\includegraphics[width=0.8\textwidth]{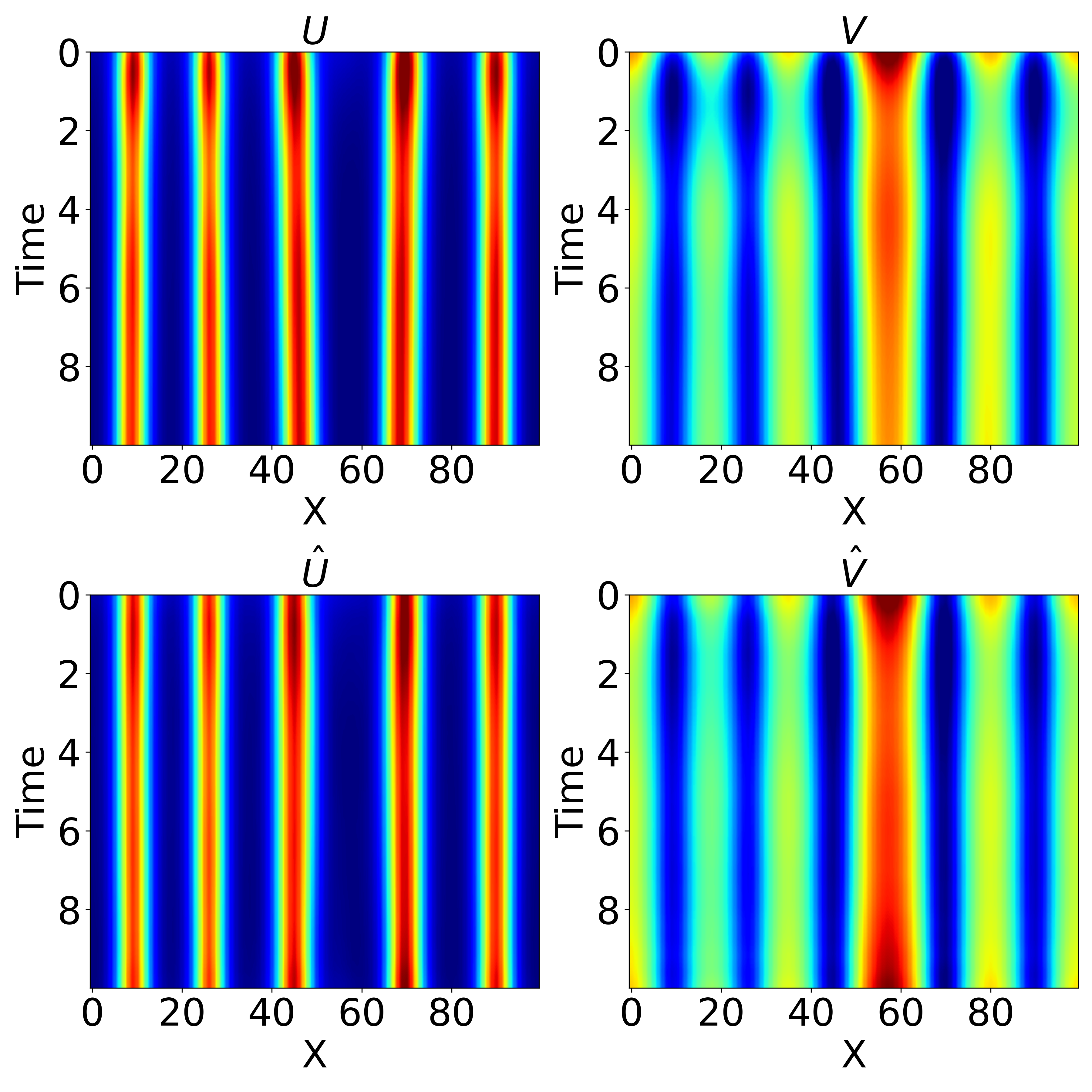}
\caption[Prediction of the 1D Turing model] {Prediction of the 1D Turing model. $U$ and $V$ are the concentrations of two diffusible substances. This model is generated on a plane space of size $100\times 1$ ($X$ represents the first dimension) and a time domain of $[0,10]$. It is parameterized using $c_1 = 0.1$, $c_2=0.9$, $c_{-1}=1$, $c_3=1$, $d_1=1$, and $d_2=40$.}
\label{fig:turing1d_prediction}
\end{figure}

Turing patterns, such as spots and stripes observed in nature, arise spontaneously and autonomously from homogeneous states through reaction-diffusion systems involving two diffusive substances that interact with each other \cite{turing1990chemical}. The 1D Turing model is a simplified version of the general Turing pattern that allows us to view the predictions over the entire time domain, and is generated on a plane space of size $100\times 1$. The Schnakenberg kinetics model \cite{maini2012turing} is used for producing this. Appendix \ref{appendix:turing} provides the complete definition and equations for this model. The prediction results for this model using SB-FNN can be seen in Figure \ref{fig:turing1d_prediction}.

\subsection{2D Turing Model}
\begin{figure}[h]
\centering
\includegraphics[width=0.8\textwidth]{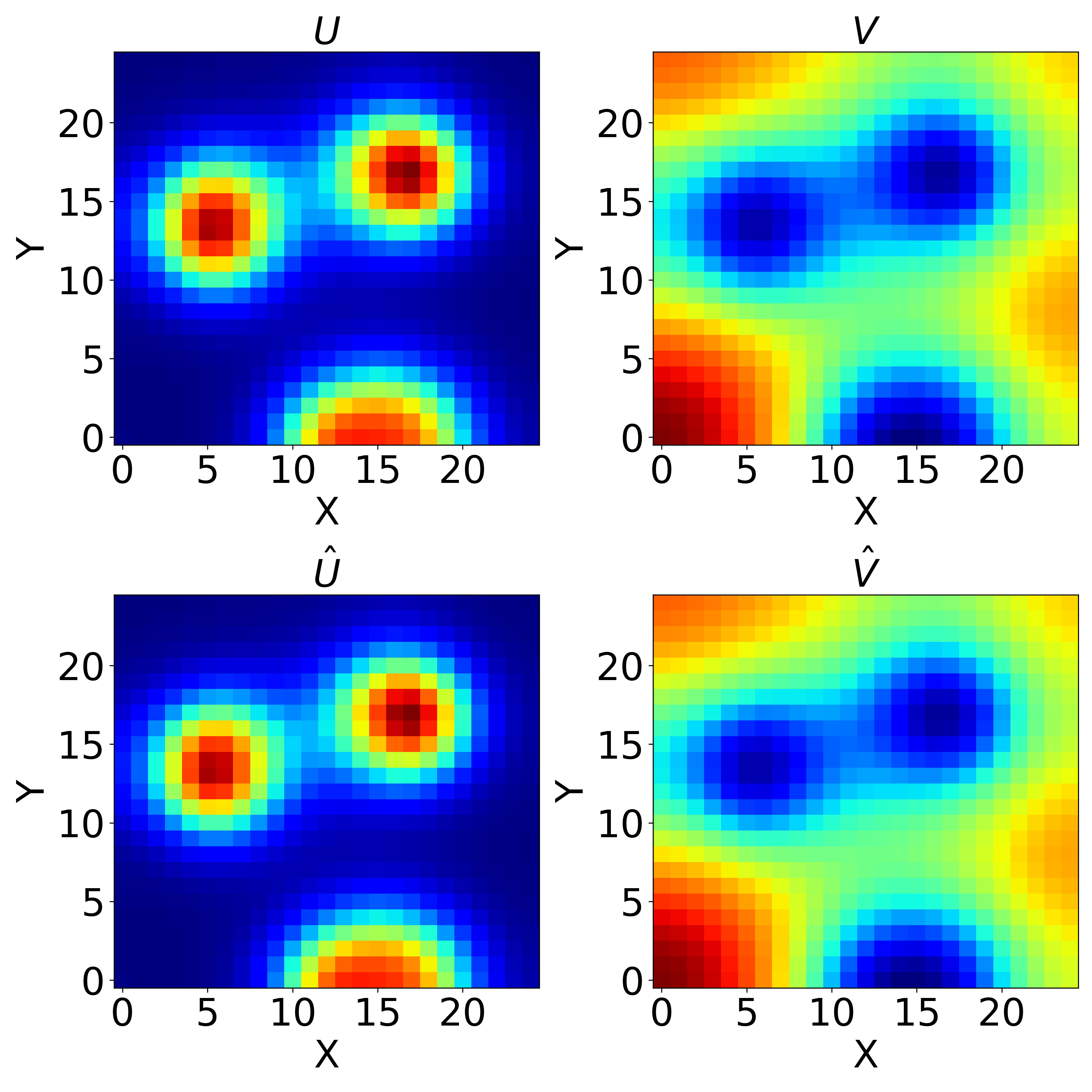}
\caption[Prediction of the 2D Turing model] {Prediction of the 2D Turing model on the last time step. $U$ and $V$ are the concentrations of two diffusible substances. This model is generated on a plane space of size $25\times 25$, and it is parameterized using $c_1 = 0.1$, $c_2=0.9$, $c_{-1}=1$, $c_3=1$, $d_1=1$, and $d_2=40$.}
\label{fig:turing2d_prediction}
\end{figure}

To focus on the reaction-diffusion patterns of Turing model on the plane, we can also produce the Turing model on a plane space of size $25\times 25$. The complete definition and equations for this model can be found in Appendix \ref{appendix:turing} as well. The prediction results for this model using SB-FNN can be seen in Figure \ref{fig:turing2d_prediction}.

\section{Performance Analysis of SB-FNN against PINN}

We assess the accuracy and efficiency performance of PINN and SB-FNN on the introduced systems biology models. 

\subsection{Accuracy Evaluation}

\begin{table*}[!htb]
\centering
\small
\caption[Performance comparison between PINN and SB-FNN on different systems biology models] {Performance comparison between PINN and SB-FNN on different systems biology models.}
\label{tab:exp1-all}
\begin{tabular}{ l cc } 
\toprule
\multirow{2}{*}{Model} & \multicolumn{2}{c}{Accuracy (N-MSE)} \\
\cmidrule(l){2-3} 
 & PINN & SB-FNN\\
\midrule 
\midrule 
{Rep3} & $1.3173 \pm 0.9355 (e^{-3})$ & $5.5949 \pm 0.3237 (e^{-5})$ \\
{Rep6} & $2.8501 \pm 0.0041 (e^{1})$ & $9.3663 \pm 1.0792 (e^{-4})$ \\
{SIR} & $6.9605 \pm 3.3973 (e^{-2})$ & $6.8226 \pm 1.0602 (e^{-5})$ \\
{A-SIR} & $1.7167 \pm 0.5161 (e^{-2})$ & $7.4510 \pm 0.6084 (e^{-5})$ \\
{1D Turing} & $1.1779 \pm 0.0901 (e^{0})$ & $1.8099 \pm 0.2314 (e^{-2})$ \\
{2D Turing} & $4.9022 \pm 0.1294 (e^{-2})$ & $4.7462 \pm 0.2989 (e^{-3})$ \\
\bottomrule
\end{tabular}
\end{table*}

\begin{figure}[h]
\centering
\includegraphics[width=0.9\textwidth]{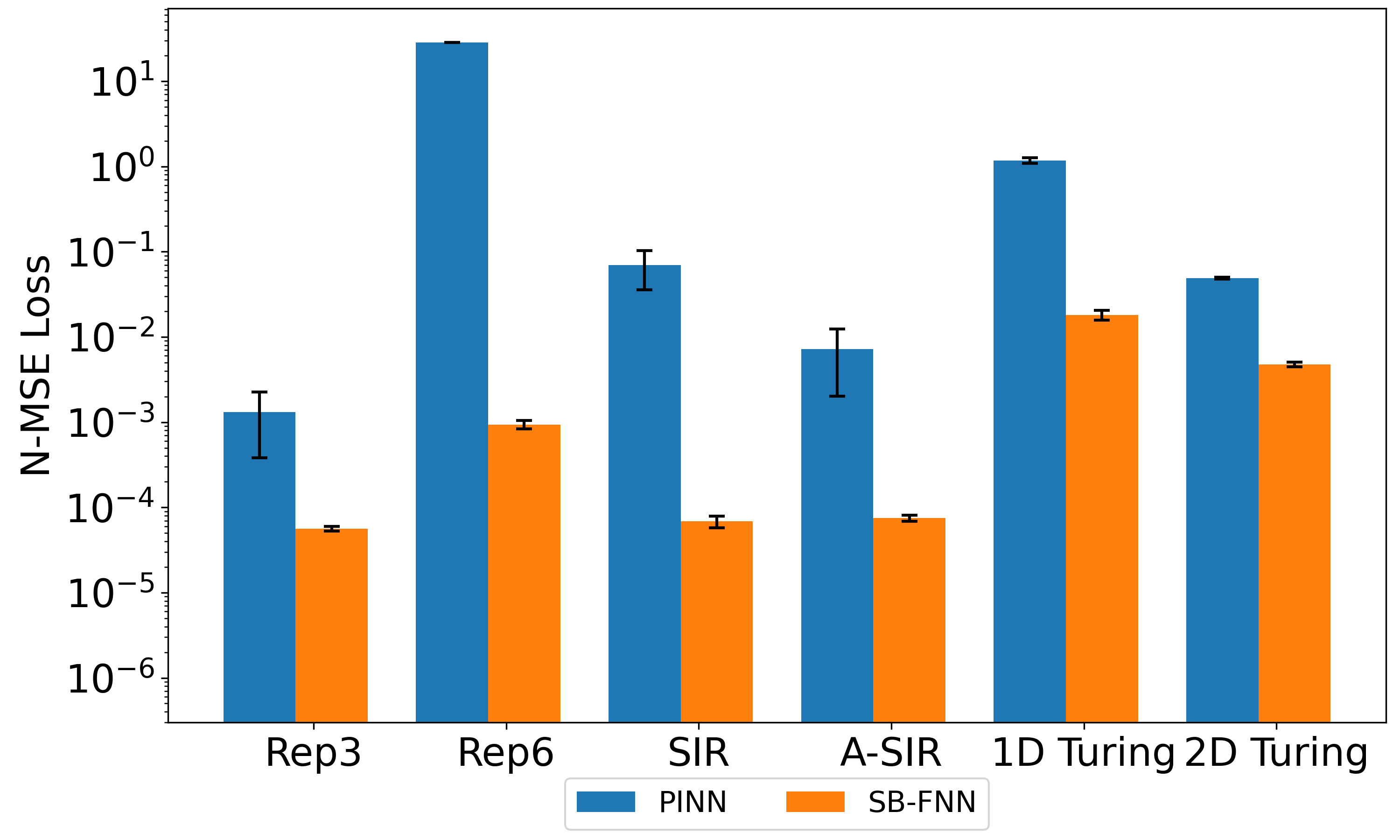}
\caption[Performance comparison between PINN and SB-FNN on different systems biology models] {Performance comparison between PINN and SB-FNN on different systems biology models. The N-MSE values between the ground truth and the predictions on the test set have been calculated and plotted. In the bar chart, the bars represent the mean N-MSE values and the error dashes indicate the standard deviation. It can be observed that PINN does not perform well in predicting the Rep6 model and the 1D Turing model, and it performs worse than SB-FNN in all the models.}
\label{fig:pinn_vs_sb_fnn}
\end{figure}

The accuracy (N-MSE) (Eq. \ref{eq:nmse}) of each model on the test set is computed by averaging the results of five experiments with randomly initialized neural networks (using different random seeds for the PyTorch, Numpy, and Random packages). 
Table \ref{tab:exp1-all} and Figure \ref{fig:pinn_vs_sb_fnn} summarize the obtained accuracy results. PINN fails in the prediction of the Rep6 model and the 1D Turing model and performs worse than SB-FNN in all models.

\begin{figure}
    \centering
    \subcaptionbox{Training loss (N-MSE) on the Rep3 model}{\includegraphics[width=0.49\textwidth]{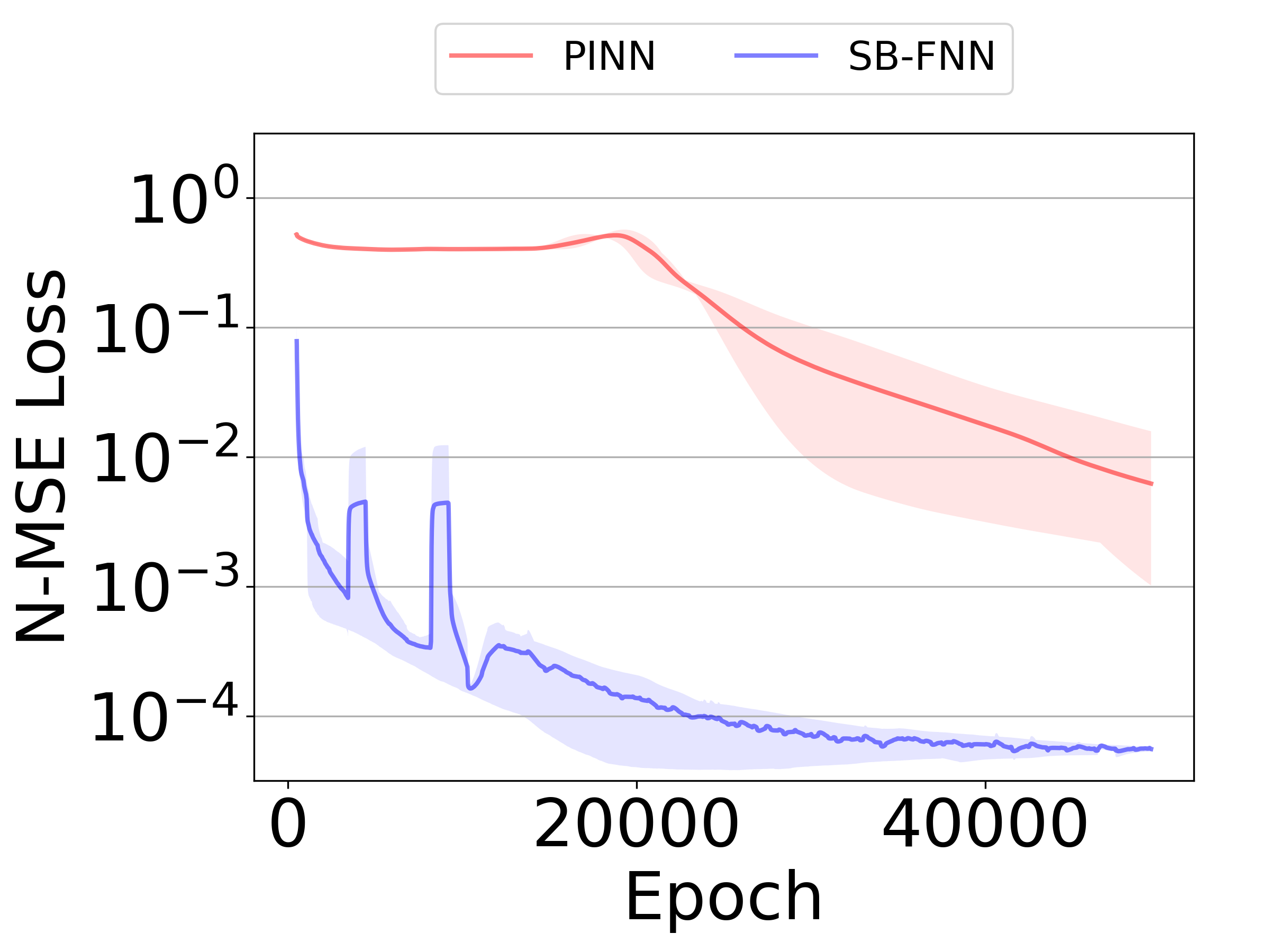}}
    \subcaptionbox{Training loss (N-MSE) on the Rep6 model}{\includegraphics[width=0.49\textwidth]{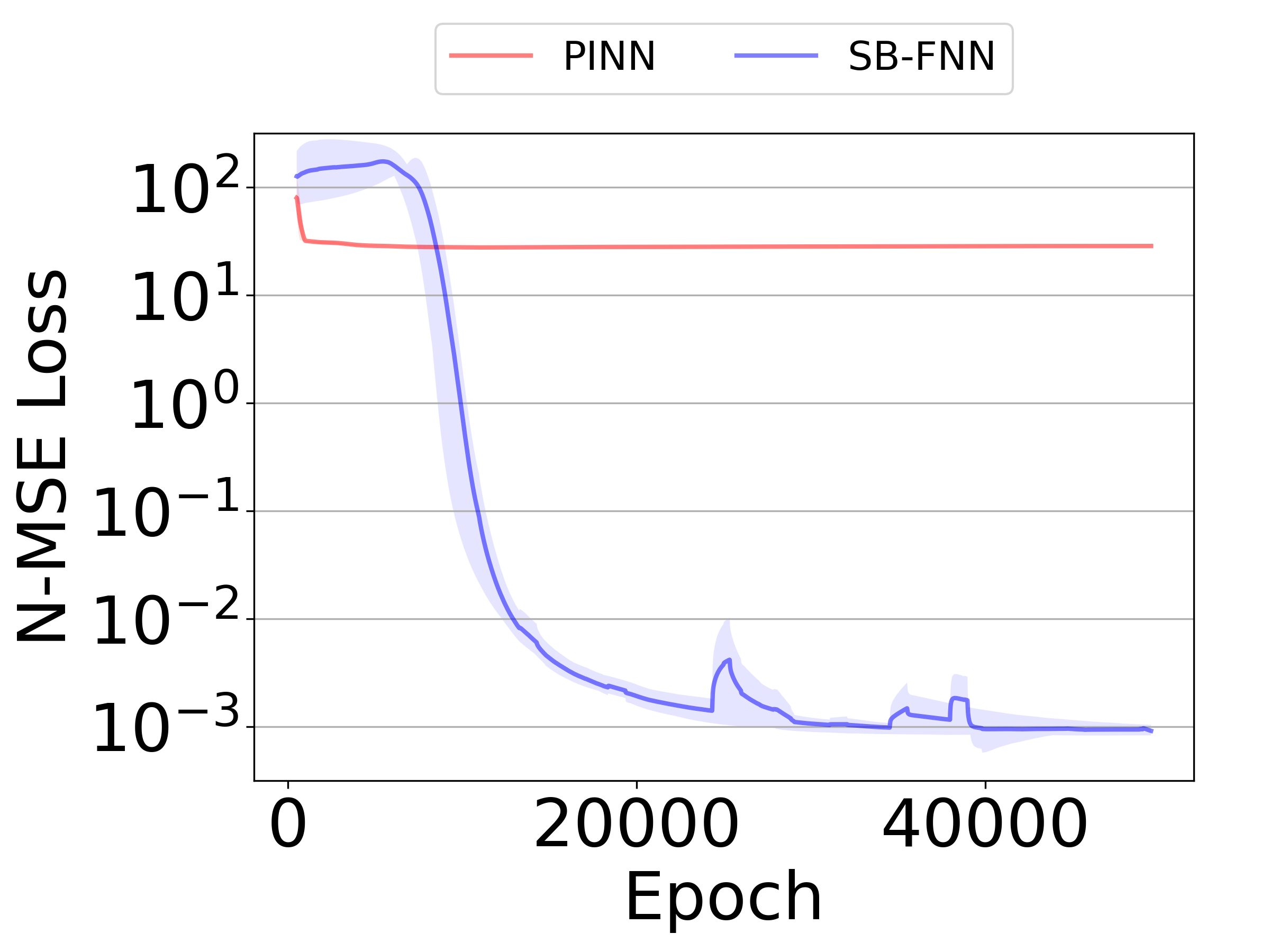}}
    \subcaptionbox{Training loss (N-MSE) on the SIR model}{\includegraphics[width=0.49\textwidth]{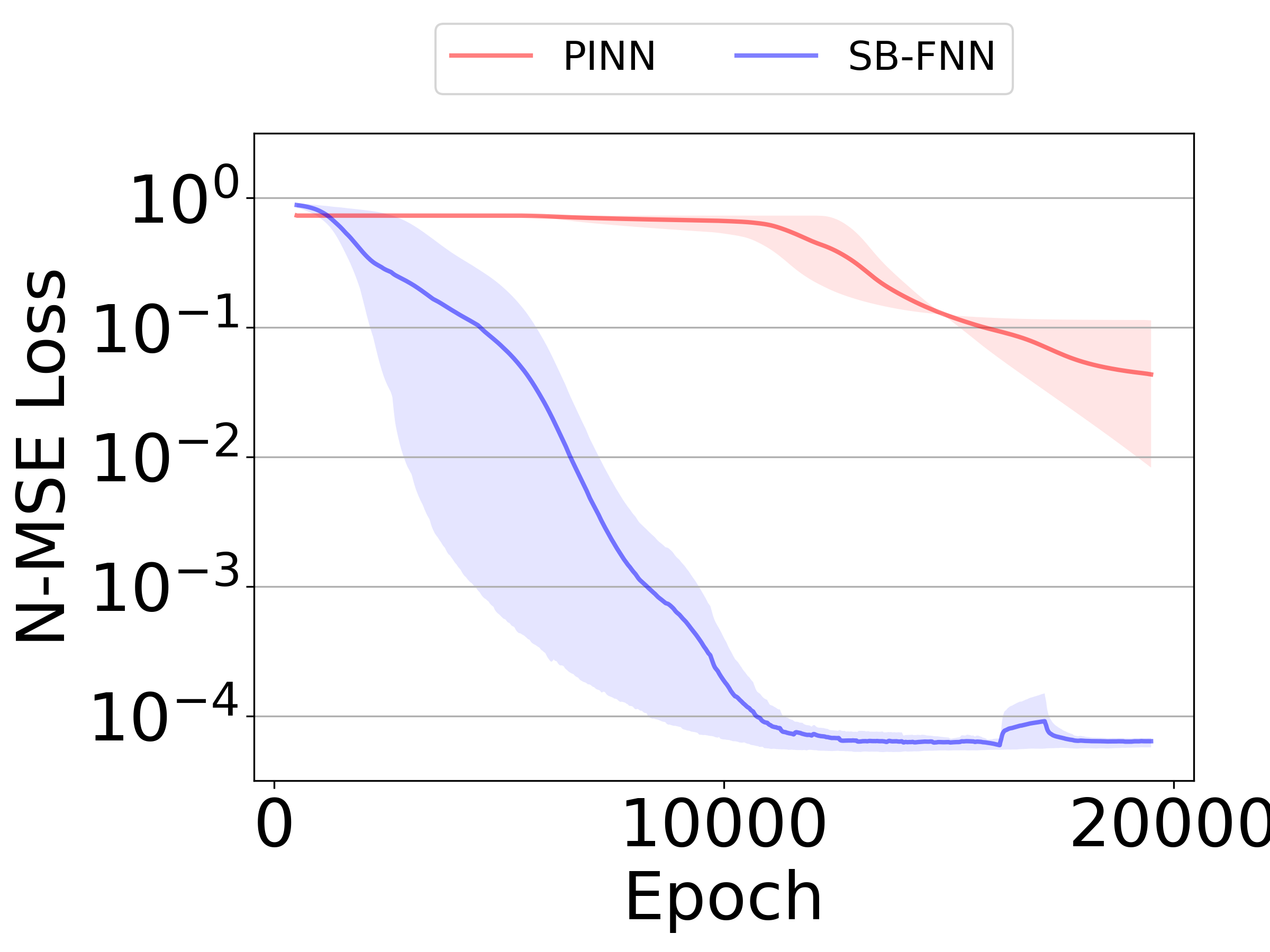}}
    \subcaptionbox{Training loss (N-MSE) on the A-SIR model}{\includegraphics[width=0.49\textwidth]{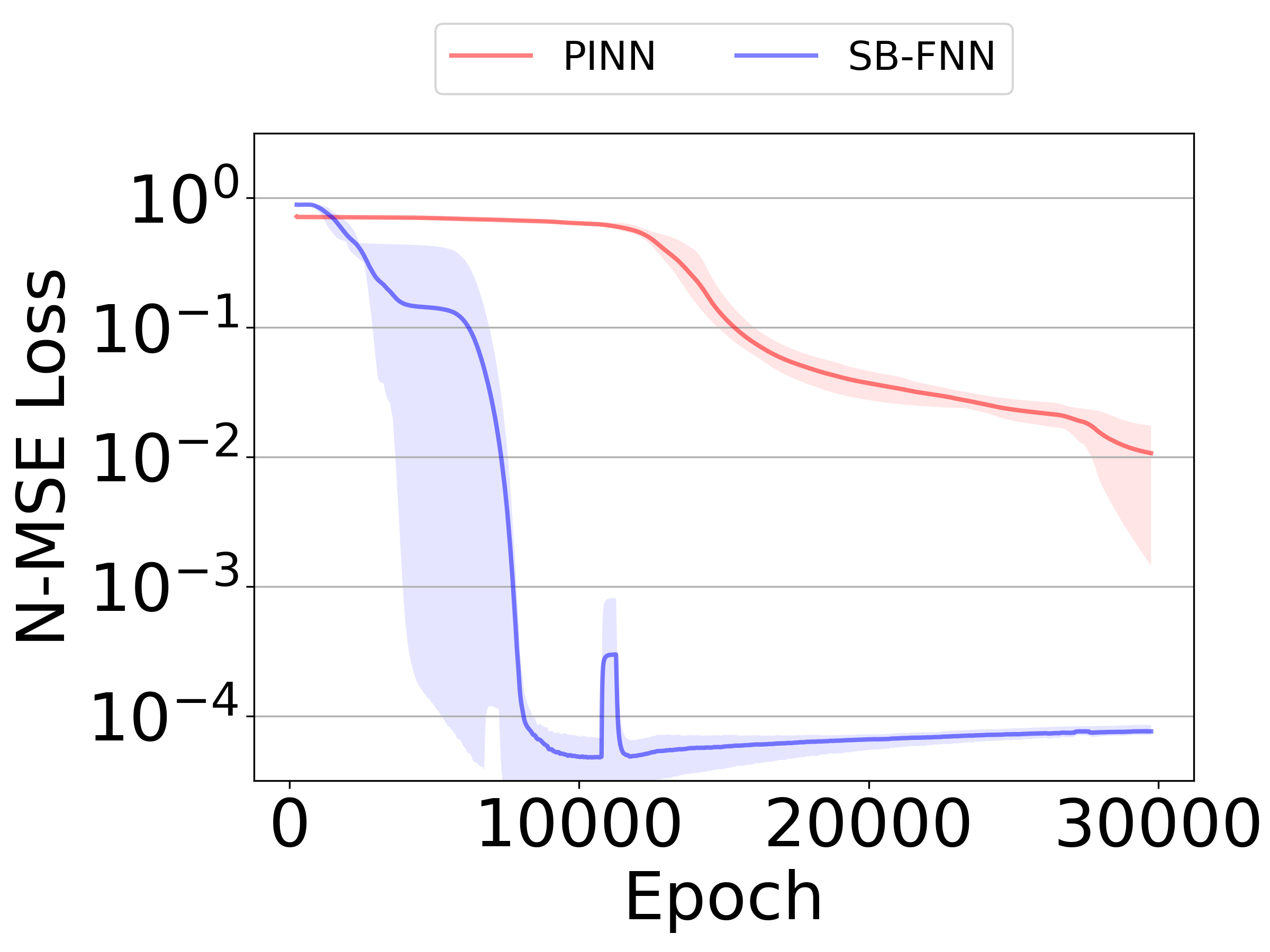}}
    \subcaptionbox{Training loss (N-MSE) on the 1D Turing model}{\includegraphics[width=0.49\textwidth]{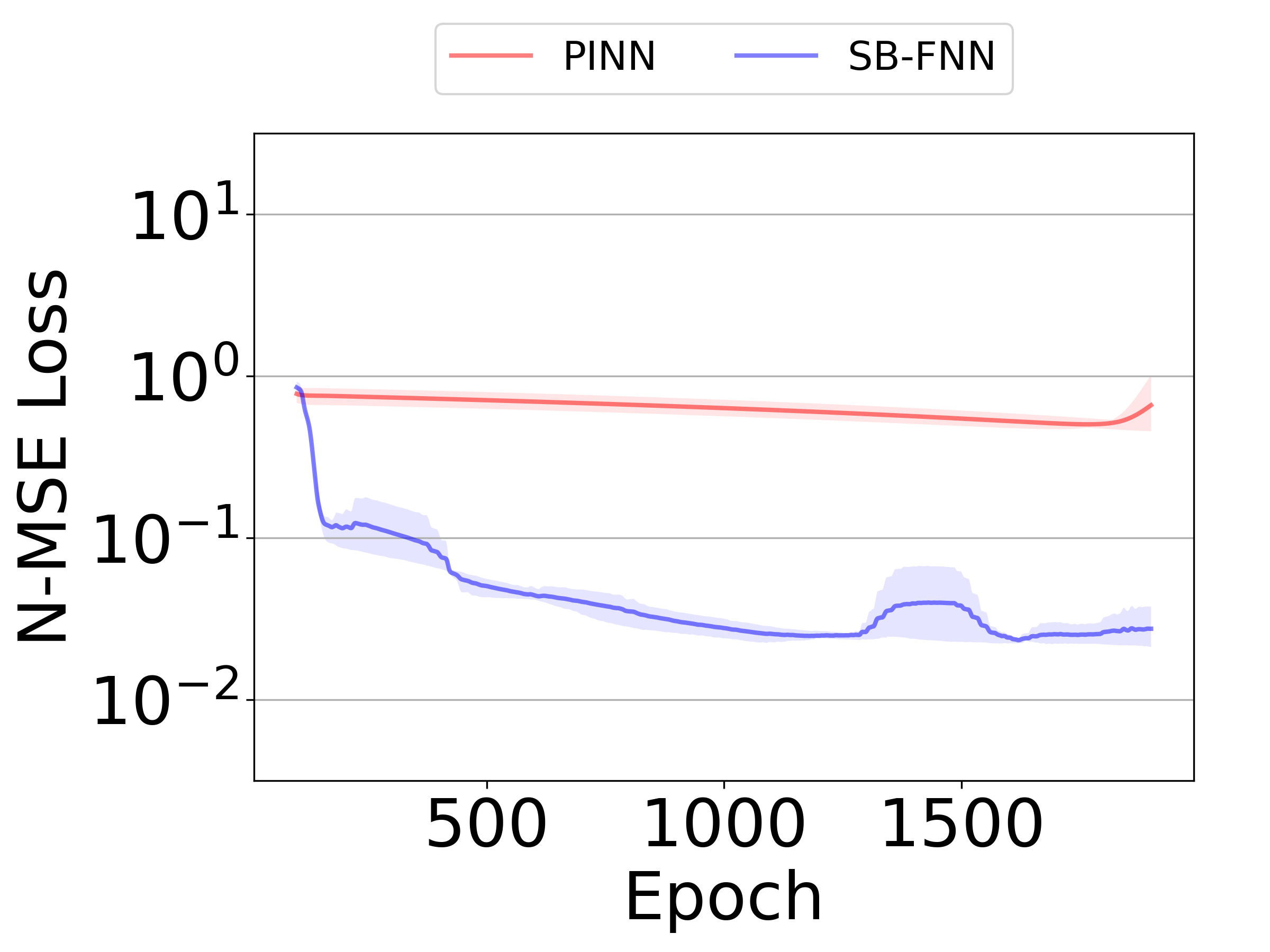}}
    \subcaptionbox{Training loss (N-MSE) on the 2D Turing model}{\includegraphics[width=0.49\textwidth]{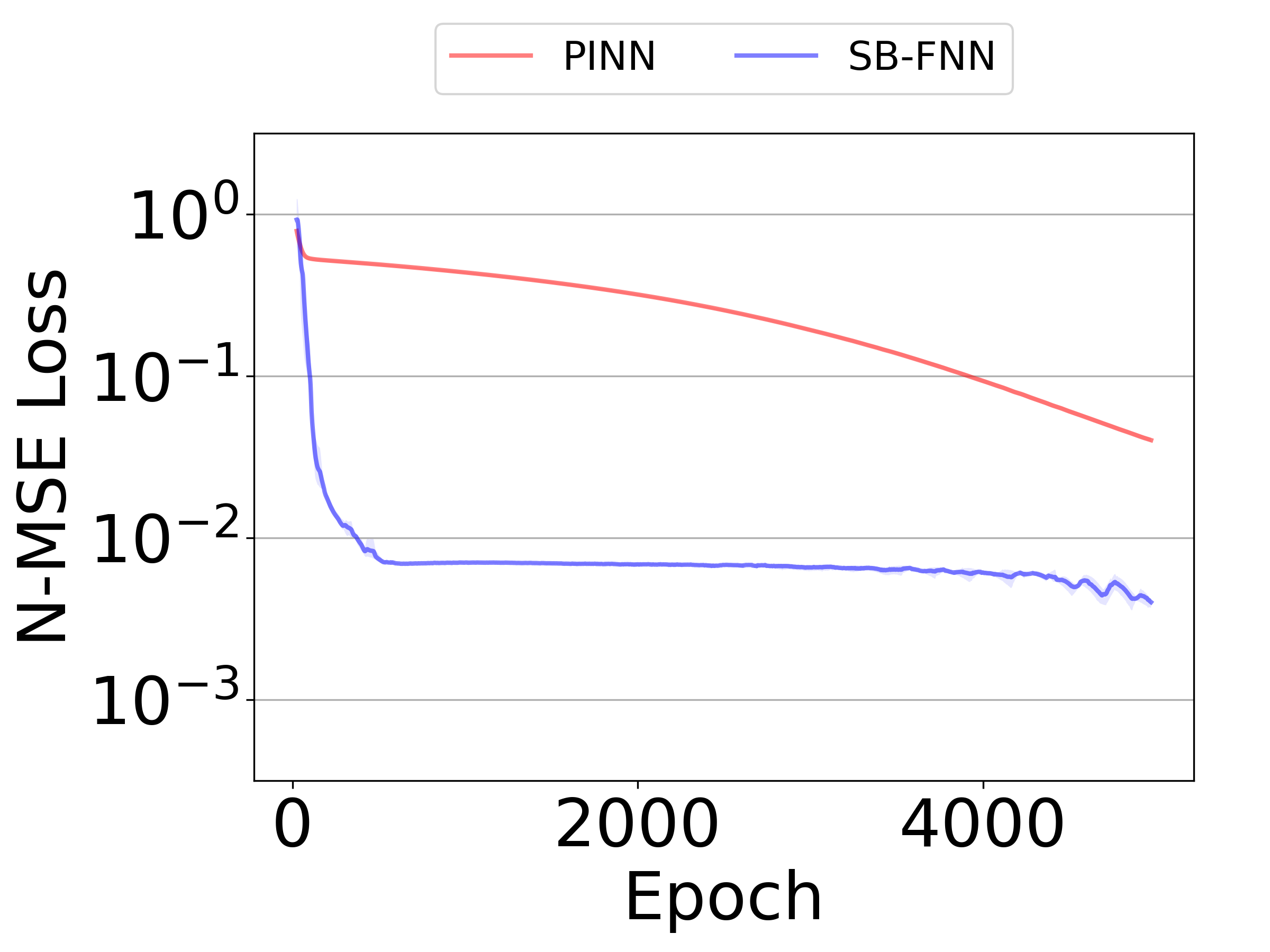}}
    \caption[Comparison of training loss (N-MSE) between PINN and SB-FNN over epochs] {Comparison of training loss (N-MSE) between PINN and SB-FNN over epochs. For each model and method used, we conduct five separate tests. The solid lines in the plots represent the average value across these tests, while the shaded areas in the corresponding lighter color show the upper and lower bounds over the tests.}
    \label{fig:pinn_vs_sb_fnn_nmse}
\end{figure}

We also plot the training loss (N-MSE) between PINN and SB-FNN over epochs, which is shown as Figure \ref{fig:pinn_vs_sb_fnn_nmse}. To ensure stable prediction results, we set the maximum epoch as follows: 50k for Rep3 and Rep6 models, 20k for the SIR model, 30k for the A-SIR model, and 5k for the 1D Turing and 2D Turing models. For each model trained with each method, the final accuracy is calculated using the average of the last 10\% of epochs, in case of unexpected small shocks in the curves.

The results presented above demonstrate that SB-FNN outperforms both PINN and FNN across all listed models. Generally, PINN shows good performance in predicting the dynamics of systems biology models with simple equations, but it performs worse in more complex models. In contrast, SB-FNN performs well in all models. These findings highlight the potential of SB-FNN in accurately predicting the dynamics of complex systems biology models.

\subsection{Efficiency Evaluation}

\begin{table*}[!htb]
\centering
\scriptsize
\caption[Efficiency evaluation of different methods] {Efficiency evaluation of different methods}
\label{tab:exp4}
\begin{tabular}{ l cccccc } 
\toprule
\multirow{2}{*}{Method} &  \multicolumn{6}{c}{Training Time per Epoch / s} \\
\cmidrule(l){2-7} 
 & Rep3 & Rep6 &  SIR & A-SIR & 1D Turing & 2D Turing \\
\midrule 
\midrule 
\multirow{2}{*}{Tanh} & $7.4903$ & $7.9043$ & $7.5411$ & $1.1423$ & $2.0235$ & $1.6177$ \\ 
 & $\pm0.6795 (e^{-2})$ & $\pm0.3299 (e^{-2})$ & $\pm0.7330 (e^{-2})$ & $\pm0.0596 (e^{-1})$ & $\pm0.1864 (e^{-1})$ & $\pm0.0327 (e^{-1})$ \\
\multirow{2}{*}{ReLU} & $6.7811$ & $7.9735$ & $7.4282$ & $1.0946$ & $2.0448$ & $1.5984$ \\ 
 & $\pm0.3456 (e^{-2})$ & $\pm0.4357 (e^{-2})$ & $\pm0.4627 (e^{-2})$ & $\pm0.0927 (e^{-1})$ & $\pm0.1665 (e^{-1})$ & $\pm0.0131 (e^{-1})$ \\
\multirow{2}{*}{Softplus} & $8.1193$ & $8.6280$ & $7.3499$ & $1.1186$ & $1.9356$ & $1.5789$ \\ 
 & $\pm0.6498 (e^{-2})$ & $\pm0.5897 (e^{-2})$ & $\pm0.3984 (e^{-2})$ & $\pm0.0955 (e^{-1})$ & $\pm0.1040 (e^{-1})$ & $\pm0.0318 (e^{-1})$ \\
\multirow{2}{*}{ELU} & $7.1769$ & $8.1191$ & $7.3310$ & $1.1580$ & $1.9357$ & $1.6085$ \\ 
 & $\pm0.2137 (e^{-2})$ & $\pm0.2983 (e^{-2})$ & $\pm0.1597 (e^{-2})$ & $\pm0.1371 (e^{-1})$ & $\pm0.2124 (e^{-1})$ & $\pm0.0292 (e^{-1})$ \\
\multirow{2}{*}{GELU} & $7.6652$ & $8.7640$ & $7.5818$ & $1.2064$ & $2.1241$ & $1.5857$ \\ 
 & $\pm1.6144 (e^{-2})$ & $\pm1.2170 (e^{-2})$ & $\pm0.7935 (e^{-2})$ & $\pm0.1172 (e^{-1})$ & $\pm0.1218 (e^{-1})$ & $\pm0.0111 (e^{-1})$ \\
\multirow{2}{*}{Sin} & $7.3184$ & $8.0606$ & $7.4377$ & $1.0406$ & $1.9541$ & $1.6388$ \\ 
 & $\pm0.3173 (e^{-2})$ & $\pm0.2456 (e^{-2})$ & $\pm0.6310 (e^{-2})$ & $\pm0.0506 (e^{-1})$ & $\pm0.1243 (e^{-1})$ & $\pm0.0245 (e^{-1})$ \\
\multirow{2}{*}{Adaptive} & $9.6632$ & $1.0841$ & $9.0508$ & $1.2492$ & $2.8061$ & $4.4034$ \\ 
 & $\pm1.2990 (e^{-2})$ & $\pm0.0428 (e^{-1})$ & $\pm0.5789 (e^{-2})$ & $\pm0.0343 (e^{-1})$ & $\pm0.1133 (e^{-1})$ & $\pm0.1157 (e^{-1})$ \\
\multirow{2}{*}{SB-FNN} & $9.0656$ & $1.0262$ & $9.0508$ & $1.2492$ & $2.8061$ & $4.4034$ \\ 
 & $\pm0.4910 (e^{-2})$ & $\pm0.0395 (e^{-1})$ & $\pm0.5789 (e^{-2})$ & $\pm0.0343 (e^{-1})$ & $\pm0.1133 (e^{-1})$ & $\pm0.1157 (e^{-1})$ \\
\multirow{2}{*}{PINN} & $7.0960$ & $8.0010$ & $6.9456$ & $1.1917$ & $3.7831$ & $1.5404$ \\ 
 & $\pm0.6624 (e^{-2})$ & $\pm0.4455 (e^{-2})$ & $\pm0.4092 (e^{-2})$ & $\pm0.0956 (e^{-1})$ & $\pm0.1542 (e^{-1})$ & $\pm0.1372 (e^{0})$ \\
 
\bottomrule
\end{tabular}%}
\end{table*}

In the experiments using different methods on various models, we have computed the average time taken per epoch, as presented in Table \ref{tab:exp4}. These experiments are conducted on the same GPU chip and operating system. Please note that since the variance constraint is not appropriate for non-oscillatory models, their ``Adaptive'' rows are the same as the ``SB-FNN'' rows. For relatively simple models, applying the adaptive function incurs about 20\% more time per epoch, and the use of variance constraint does not result in a significant difference in time cost. On the other hand, the performance of the PINN method in terms of time is similar to that of single activation methods on simple models, but the time cost rises sharply as the model becomes more complex. In the 2D Turing model, PINN takes more than three times as much time as SB-FNN.

\section{Ablation Analysis: Adaptive Activation Function}

To evaluate the effectiveness of the proposed adaptive activation function, we conducted a comparative analysis with six commonly used activation functions: GELU, tanh, ReLU, sin, ELU, and Softplus. We vary the activation function and calculated the accuracy (N-MSE) on the same test sets for each model, as presented in Table \ref{tab:exp2}.

\begin{table*}[!htb]
\centering
\scriptsize
\caption[Accuracy comparison of activation functions on different models] {Accuracy comparison of activation functions on different models}
\label{tab:exp2}
\begin{tabular}{ l cccccc } 
\toprule
\multirow{2}{*}{Activation} & 
\multicolumn{6}{c}{Accuracy (N-MSE)} \\
\cmidrule(l){2-7} 
Function & Rep3 & Rep6 &  SIR & A-SIR & 1D Turing & 2D Turing \\
\midrule 
\midrule 
\multirow{2}{*}{Tanh} & $5.4551$ & $9.4097$ & $1.1149$ & $1.3860$ & $2.1986$ & $6.9811$ \\ 
 & $\pm0.4950 (e^{-5})$ & $\pm0.3272 (e^{-4})$ & $\pm1.2067 (e^{-1})$ & $\pm0.0297 (e^{-4})$ & $\pm0.0616 (e^{-2})$ & $\pm0.1095 (e^{-3})$ \\
\multirow{2}{*}{ReLU} & $6.3463$ & $1.1152$ & $2.3020$ & $3.2724$ & $2.7842$ & $7.0822$ \\ 
 & $\pm4.2477 (e^{-4})$ & $\pm1.1855 (e^{2})$ & $\pm1.1113 (e^{-1})$ & $\pm0.6692 (e^{-1})$ & $\pm3.0169 (e^{-1})$ & $\pm0.0343 (e^{-3})$ \\
\multirow{2}{*}{Softplus} & $2.3713$ & $1.3282$ & $2.0404$ & $9.6764$ & $7.5665$ & $5.9022$ \\ 
 & $\pm1.4591 (e^{-4})$ & $\pm0.1320 (e^{-3})$ & $\pm0.5914 (e^{-4})$ & $\pm0.5277 (e^{-5})$ & $\pm5.6766 (e^{-2})$ & $\pm0.4148 (e^{-3})$ \\
\multirow{2}{*}{ELU} & $1.2048$ & $6.5232$ & $2.2249$ & $8.3467$ & $1.4560$ & $5.0241$ \\ 
 & $\pm0.1063 (e^{-4})$ & $\pm9.2245 (e^{1})$ & $\pm0.6665 (e^{-4})$ & $\pm1.2210 (e^{-5})$ & $\pm2.0326 (e^{0})$ & $\pm1.5024 (e^{-3})$ \\
\multirow{2}{*}{GELU} & $1.0203$ & $2.0271$ & $1.1022$ & $7.0913$ & $1.8851$ & $4.3966$ \\ 
 & $\pm0.5524 (e^{-4})$ & $\pm0.6217 (e^{-3})$ & $\pm0.1666 (e^{-4})$ & $\pm1.0441 (e^{-5})$ & $\pm0.1120 (e^{-2})$ & $\pm0.0923 (e^{-3})$ \\
\multirow{2}{*}{Sin} & $3.9671$ & $1.4455$ & $7.4375$ & $5.8169$ & $5.8022$ & $4.5739$ \\ 
 & $\pm0.4856 (e^{-5})$ & $\pm0.5674 (e^{-3})$ & $\pm0.3368 (e^{-1})$ & $\pm1.0717 (e^{-1})$ & $\pm1.4546 (e^{-2})$ & $\pm1.5236 (e^{-3})$ \\
% \multirow{2}{*}{Adaptive} & $6.8001$ & $1.0955$ & $6.8226$ & $7.4510$ & $1.8099$ & $4.7462$ \\ 
%  & $\pm2.4594 (e^{-5})$ & $\pm0.0484 (e^{-3})$ & $\pm1.0602 (e^{-5})$ & $\pm0.6084 (e^{-5})$ & $\pm0.2314 (e^{-2})$ & $\pm0.2989 (e^{-3})$ \\
% \multirow{2}{*}{Adaptive} & $6.8001$ & $1.0955$ & $6.8226$ & $7.4510$ & $1.8099$ & $4.7462$ \\ 
%  & $\pm2.4594 (e^{-5})$ & $\pm0.0484 (e^{-3})$ & $\pm1.0602 (e^{-5})$ & $\pm0.6084 (e^{-5})$ & $\pm0.2314 (e^{-2})$ & $\pm0.2989 (e^{-3})$ \\
 \multirow{2}{*}{\textbf{Adaptive}} & \textbf{6.8001} & \textbf{1.0955} & \textbf{6.8226} & \textbf{7.4510} & \textbf{1.8099} & \textbf{4.7462} \\
& \textbf{$\pm$2.4594(e$^{-5}$)} & \textbf{$\pm$0.0484(e$^{-3}$)} & \textbf{$\pm$1.0602(e$^{-5}$)} & \textbf{$\pm$0.6084(e$^{-5}$)} & \textbf{$\pm$0.2314(e$^{-2}$)} & \textbf{$\pm$0.2989(e$^{-3}$)} \\
 
\bottomrule
\end{tabular}%}
\end{table*}

Based on Table \ref{tab:exp2}, we can see that no activation function performs the best across all models. To analyze the overall performance of each activation function, we calculate an average ranking score for each one. For each of the six models, we rank the seven activation functions from 1 to 7 based on their accuracy performance (lower N-MSE value indicates a better performance). We then rank them and calculate the mean  score for each activation function over all six models.

\begin{table*}[!htb]
\centering
\small
\caption[Scoring of activation functions on different models] {Scoring of activation functions on different models}
\label{tab:exp2-2}
\begin{tabular}{ l cccccc | c } 
\toprule
\multirow{2}{*}{Activation} & 
\multicolumn{7}{c}{Score} \\
\cmidrule(l){2-8} 
Function & Rep3 & Rep6 &  SIR & A-SIR & 1D Turing & 2D Turing & mean \\
\midrule 
\midrule 

Tanh & 6 & 7 & 3 & 3 & 5 & 2 & 4.33 \\
ReLU & 1 & 1 & 2 & 2 & 2 & 1 & 1.50 \\
Softplus & 2 & 5 & 5 & 4 & 3 & 3 & 3.67 \\
ELU & 3 & 2 & 4 & 5 & 1 & 4 & 3.17 \\
GELU & 4 & 3 & 6 & 7 & 6 & 7 & 5.50 \\
Sin & 7 & 4 & 1 & 1 & 4 & 6 & 3.83 \\
% Adaptive & 5 & 6 & 7 & 6 & 7 & 5 & 6.00 \\ 
\textbf{Adaptive} & \textbf{5} & \textbf{6} & \textbf{7} & \textbf{6} & \textbf{7} & \textbf{5} & \textbf{6.00} \\  

\bottomrule
\end{tabular}%}
\end{table*}

\begin{figure}[h]
\centering
\includegraphics[width=0.9\textwidth]{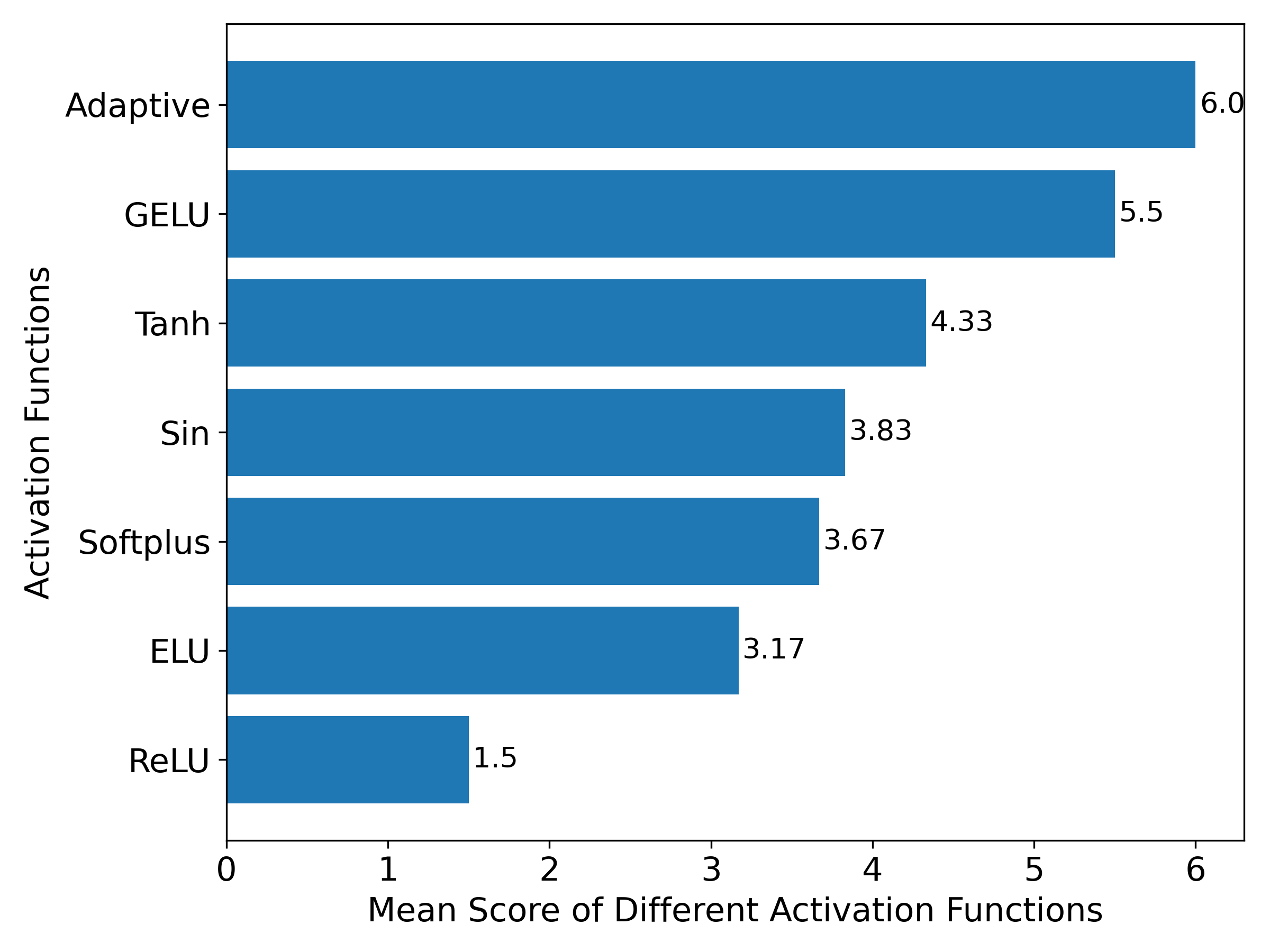}
\caption[Mean score of different activation functions in descending order.] {Mean score of different activation functions in descending order. For each of the six models, we ranked the seven activation functions based on their accuracy performance using a scoring system in which a lower N-MSE value indicates better performance and receives a higher score. We then calculated the mean score for each activation function and order them. This plot shows that although the adaptive activation function does not perform the best among all the models, it has achieved the best overall performance in terms of prediction accuracy.}
\label{fig:rank_score}
\end{figure}

Table \ref{tab:exp2-2} displays the accuracy performance comparison of methods using various activation functions. The lower N-MSE value in Table \ref{tab:exp2} indicates a better performance, which corresponds to a higher rank score. Additionally, the mean rank score in descending order is shown in Figure \ref{fig:rank_score}. The results indicate that the adaptive activation function performs the best among the six models tested, followed by GELU and Tanh. The adaptive strategy in the activation function allows the model to dynamically select the most suitable weights of activation function for different types of system biology models. Therefore, it is concluded that the adaptive activation function is the most effective among the tested models for this task.

\section{Ablation Analysis: Variance Constraint for Oscillatory Systems}

% System biology models often exhibit ubiquitous and important oscillation patterns. Here we apply the penalty function proposed in \ref{cha:cyclic} the two oscillatory models we talked about: Repressilator: Protein only and Repressilator: Protein and mRNA. We perform ablation experiments on whether adding this penalty part into the loss function or not.
System biology models often exhibit ubiquitous and important oscillation patterns. Therefore, we apply the variance constraint proposed in \ref{cha:cyclic} to the SB-FNN to predict the two oscillatory models, Repressilator: protein only model and Repressilator: mRNA and protein model. We conduct ablation experiments to evaluate the impact of adding this penalty part into the loss function.

\begin{table*}[!htb]
\centering
\footnotesize
\caption[Accuracy comparison with and without variance constraint] {Accuracy comparison with and without variance constraint.}
\label{tab:exp3}
\begin{tabular}{ l cc } 
\toprule
\multirow{2}{*}{Method} & 
\multicolumn{2}{c}{Accuracy (N-MSE)} \\
\cmidrule(l){2-3} 
 & Rep3 & Rep6 \\
\midrule 
\midrule
SB-FNN (GELU) & $1.0203 \pm 0.5524 (e^{-4})$ & $2.0271 \pm 0.6217 (e^{-3})$ \\
SB-FNN (GELU + Constraint) & $6.8936 \pm 0.5735 (e^{-5})$ & $1.1477 \pm 0.1066 (e^{-3})$ \\
SB-FNN (Adaptive) & $6.8001 \pm 2.4594 (e^{-5})$ & $1.0955 \pm 0.0484 (e^{-3})$ \\
SB-FNN (Adaptive + Constraint) & $5.5949 \pm 0.3237 (e^{-5})$ & $9.3663 \pm 1.0792 (e^{-4})$ \\

\bottomrule
\end{tabular}%}
\end{table*}

\begin{figure}[h]
\centering
\includegraphics[width=0.9\textwidth]{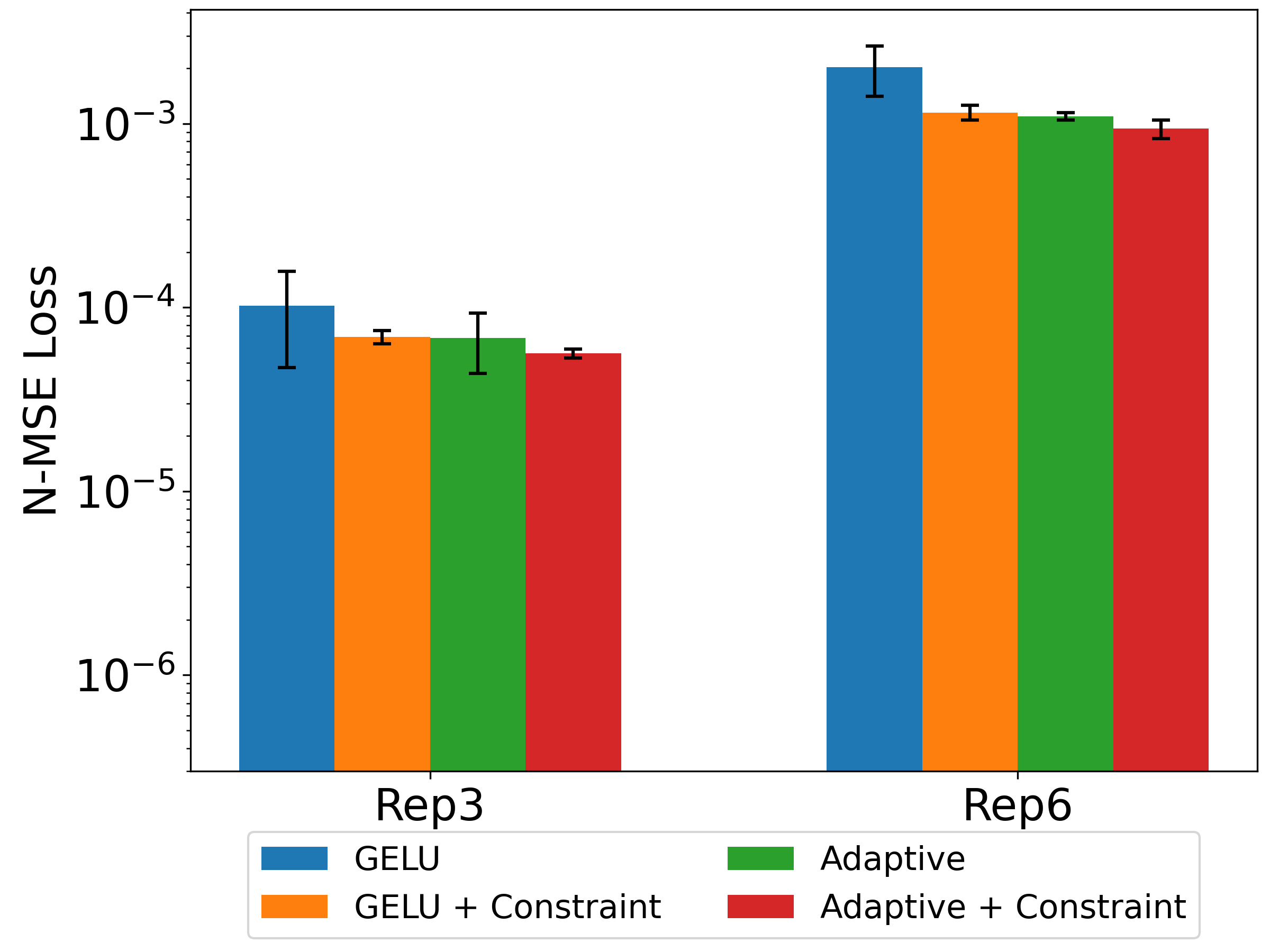}
\caption[Accuracy comparison of methods with and without variance constraint] {Accuracy comparison of methods with and without variance constraint. }
\label{fig:variance}
\end{figure}

Table \ref{tab:exp3} and Figure \ref{fig:variance} displays a comparison of the accuracy performance for methods with and without variance constraint. The methods used in the experiment were based on GELU and Adaptive activation functions, with the same starting learning rate of $0.01$ and epoch of $50$k. Except for the adaptive activation function, we take GELU into this experiment because it is designed as the original activation function applied to the FNN, and FNN is a base part of our SB-FNN. The N-MSE value was averaged over five tests with different random seeds for predicting the dynamics of two oscillatory models. The results of the ablation experiment demonstrate the effectiveness of the variance constraint in improving the performance of FNN-based models on predicting oscillatory dynamics.

\chapter{Conclusion}
In this context, the proposed Fourier-enhanced Neural Networks for systems biology (SB-FNN) is an alternative approach inspired by PINN that can handle complex biological models more efficiently and accurately. SB-FNN has an embedded Fourier neural network with a customized design that incorporates two features for systems biology applications: an adaptive activation function and a variance constraint.

The embedded Fourier neural network, which is a type of neural network that represents the input data in the frequency domain using Fourier transforms, enables the model to capture complex patterns and features that may not be easily detectable in the spatial domain. This capability is particularly valuable in systems biology, where many biological processes exhibit oscillatory patterns or other complex temporal dynamics. By representing the input data in the frequency domain, SB-FNN can effectively capture these dynamics, resulting in more accurate predictions.

The adaptive activation function for each Fourier layer, replacing the original GeLU, offers a promising avenue for improving the performance and interpretability of deep learning models for physical systems. By leveraging prior knowledge about the system and carefully selecting the appropriate set of activation functions, we can build models that are both more effective and adaptable to different systems biology models. Moreover, the use of adaptive activation functions makes it easier to interpret the model's behavior and provides a more intuitive understanding of the underlying biological processes.

The variance constraint focuses on the oscillatory pattern of some system biology models. It helps to incorporate the periodicity constraint into the loss function, further improving the accuracy of SB-FNN on the models involving oscillatory patterns. This is particularly important in systems biology, where many biological processes exhibit periodic behavior, such as the circadian rhythm or cell cycle.

One limitation of this study is that it does not perform well on predicting some stiff dynamics produced by differential equations, which are commonly seen in some Cell-cycle models. One potential solution is to apply some wavelets method which can represent PDE dynamics as a linear combination of wavelet basis functions. Except for applying new base level techniques, in the future we can also explore the inverse techniques based on SB-FNN to estimate the best mapping parameters for structure-specified models using given dynamic truth.

In summary, SB-FNN offers a promising method for efficiently and accurately predicting the dynamics of complex systems biology models. By incorporating the embedded Fourier neural network, adaptive activation functions, and variance constraint, SB-FNN is particularly well-suited to handling biological systems, especially those exhibiting ubiquitous and vital oscillatory patterns. While further research is necessary to determine the full potential of SB-FNN, this method represents an exciting avenue for integrating machine learning and mathematical modeling to make new discoveries in biology and health.

\bibliographystyle{unsrt}  
\bibliography{bibliography}

\appendix
\chapter{Systems Biology Models}
\label{appendix:models}
Systems biology models are computational tools used to study complex biological systems. These models can help us understand the behavior of biological systems, predict their responses to various stimuli, and design interventions to alter their behavior. In this appendix, we will introduce some popular types of systems biology models.

% \section{Predator–Prey}
% \label{sup:pp}

% The predator-prey model describes the dynamics of two interacting species, predators and prey. The population change over time is governed by a pair of nonlinear differential equations:
% \begin{equation}
% \begin{cases}
% \dfrac{dU}{dt} = \alpha U - \beta UV \vspace{1ex} \\
% \dfrac{dV}{dt} = - \gamma V + \delta UV \vspace{1ex} 
% \end{cases}
% \label{eq:pp}
% \end{equation}
% In Eqs.~\ref{eq:pp}, $U$ and $V$ represent the population densities of prey and predators. Parameters $\alpha, \beta, \gamma, \delta$ are the rate constants for the birth of prey, killing because of predators, death of predators, and reproduction because of prey, respectively.

% \subsubsection{Lorenz Model}
% \label{sup:lorenz}
% The Lorenz system \cite{lorenz1962statistical} is a system of ordinary differential equations first studied by mathematician and meteorologist Edward Lorenz. The model is a system of three ordinary differential equations:

% \begin{equation}
% \begin{cases}
% \dfrac{dX}{dt} = \sigma\left(Y-X\right) \vspace{1ex} \\
% \dfrac{dY}{dt} = X\left(\rho-Z\right)-Y \vspace{1ex} \\
% \dfrac{dZ}{dt} = XY-\beta Z\vspace{1ex} 
% \end{cases}
% \label{eq:lorenz}
% \end{equation}

% In equation \ref{eq:lorenz}, $X$ is proportional to the rate of convection, $Y$ to the horizontal temperature variation, and $Z$ to the vertical temperature variation \cite{sparrow2012lorenz}. The constants $\sigma$, $\rho$, and $\beta$ are system parameters proportional to the Prandtl number, Rayleigh number, and certain physical dimensions of the layer itself.

\section{Repressilator Model: mRNA and Protein (Rep6)}
\label{appendix:rep6}
The repressilator model \cite{elowitz2000synthetic} describes a synthetic oscillatory system of transcriptional repressors. 
The transcription and translation reactions among three pairs of repressor mRNAs and proteins can be represented by the following kinetic equations: 
\begin{equation}
\begin{cases}
\dfrac{dM_{i}}{dt} = - M_{i}+\dfrac{\alpha}{1+P^{n}_{j}}+\alpha_{0}\quad\vspace{1ex} \\
\dfrac{dP_{i}}{dt} = - \beta\left(P_{i}-M_{i}\right)\quad\vspace{1ex} 
\end{cases}
\begin{pmatrix} i=lacI, tetR, cI \\ j=cI, lacI, tetR \end{pmatrix}
\label{eq:repressilator6}
\end{equation}
In Eqs.~\ref{eq:repressilator6}, $P_{i}$ denotes the repressor-protein concentrations, and $M_{i}$ represents the corresponding mRNA concentrations, where $i$ is $lacI$, $tetR$, or $cI$. If there are saturating amounts of repressor, the number of protein copies produced from a given promoter type is $\alpha_{0}$. Otherwise, this number would be $\alpha+\alpha_{0}$ per cell. $\beta$ represents the ratio of the protein decay rate to the mRNA decay rate, and $n$ is a Hill coefficient.

\section{Repressilator: Protein Only (Rep3)}
\label{appendix:rep3}
If we focus exclusively on the proteins, we can simplify the Repressilator model, which includes both mRNA and protein (Rep6), to a protein-only version (Rep3) as shown below:
\begin{equation}
% \begin{cases}
\dfrac{dP_{i}}{dt} = \dfrac{\beta}{1+P^{n}_{j}} - P_{i}\quad\vspace{1ex} 
% \end{cases}
\begin{pmatrix} i=lacI, tetR, cI \\ j=cI, lacI, tetR \end{pmatrix}
\label{eq:repressilator3}
\end{equation}
The variables utilized in Eqs.~\ref{eq:repressilator3} have the same connotations as those in Eqs.~\ref{eq:repressilator6}.

\section{SIR Model}
\label{appendix:sir}
The SIR \cite{anderson1991discussion} is a classic compartmental model in epidemiology to simulate the spread of infectious disease. The basic SIR model considers a closed population with three different labels susceptible (S), infectious (I), and recovered (R).
The evolution of the three interacting groups is predicted by the following equations:
\begin{equation}
\begin{cases}
\dfrac{dS}{dt} = -\dfrac{\beta IS}{N} \vspace{1ex} \\
\dfrac{dI}{dt} = \dfrac{\beta IS}{N} - \gamma I\vspace{1ex} \\
\dfrac{dR}{dt} = \gamma I \vspace{1ex}
\end{cases}
\label{eq:sir}
\end{equation}
where $S$, $I$, and $R$ represent the susceptible, infected, and remove (either by death or recovery) populations, respectively. $\beta$ refers to the contact rate between the susceptible and infected individuals, and $\gamma$ refers to the removal rate of the infected population.

\section{Age-structured SIR Model (A-SIR)}
\label{appendix:asir}
Many infectious diseases, such as COVID-19, have dramatically different effects on individuals of different ages. 
According to this, an age-structured SIR model was proposed to take into consideration the age-group difference~\cite{ram2021modified}. 
In the age-structured SIR (ASIR) model (see below Eqs.~\ref{eq:sir-ages}), the functions $S_i(t)$, $I_i(t)$, and $R_i(t)$ represent the susceptible, infectious, and removed population at the $i$th age-bracket for $1\leq i \leq n$, where $n$ is the number of age-groups. $\mathcal{M}\in \mathbb{R}^{n\times n}$ is an age-contact matrix describing the rate of contact between each pair of age-groups, and the value used comes from~\cite{ram2021modified}.
\begin{equation}
\begin{cases}
\dfrac{dS_{i}}{dt} = - \dfrac{\beta S_{i}}{N}\cdot \sum ^{n}_{j=1}\mathcal{M}_{ij} I_{j}\vspace{1ex} \\
\dfrac{dI_{i}}{dt} =  \dfrac{\beta S_{i}}{N}\cdot \sum ^{n}_{j=1}\mathcal{M}_{ij} I_{j} - \gamma I_{i}\vspace{1ex} \\
\dfrac{dR_{i}}{dt} = \gamma I_{i} \vspace{1ex}
\end{cases}
\label{eq:sir-ages}
\end{equation}

% \begin{figure}[h]
% \centering
% \includegraphics[width=0.4\textwidth]{fig/SIR_Ages_matrix.png}
% \caption{An example for the age-contact matrix $\mathcal{M}$ in the age-structured SIR model.}
% \label{fig:sirages_matrix}
% \end{figure}

\section{Turing Model (1D and 2D)}
\label{appendix:turing}
Turing patterns, such as spots and stripes observed in nature, arise spontaneously and autonomously from initially uniform states and can be mathematically described by reaction-diffusion systems involving two interacting and diffusive substances~\cite{turing1990chemical}. These models can exhibit substantial complexity and are highly dynamic. In this study, we focus on a two-dimensional Turing model known as the Schnakenberg kinetics \cite{maini2012turing}:
% maini2012turing  Schnakenberg model https://core.ac.uk/download/pdf/8791221.pdf
\begin{equation}
\begin{cases}
\dfrac{\partial U}{\partial t}=c_{1}-c_{\text{-}1}U+c_{3}U^{2}V+d_{1}\nabla^{2} U \vspace{1ex}\\
\dfrac{\partial V}{\partial t}=c_{2}-c_{3}u^{2}v +d_{2}\nabla^{2} V \vspace{1ex}
\end{cases}
\end{equation}
where $U$ and $V$ are the concentrations of two diffusible substances, $c_{\text{-}1}, c_{1}, c_{3}$ represent the deterministic reaction rates, $d_{1}$, $d_{2}$ are diffusion rates, and the rest are reaction terms for the two substances.

The Turing model is commonly visualized in two dimensions (2D Turing), but it can be challenging to display 2D predictions for the entire time domain. To overcome this challenge, we limit the model to one dimension, so that at each time step within the time domain, a prediction can be made in one dimension. The predictions can then be arranged along the time domain to form a figure, which is named as the 1D Turing model.

% \vita
% \chapter{}

% % \chapter{Curriculum Vitae}
% \thispagestyle{empty}
% \begin{center}
% \includegraphics[page=1, width=\textwidth]{CV_EnzeXu.pdf}
% \end{center}
% \clearpage

% \thispagestyle{empty}
% \begin{center}
% \includegraphics[page=2, width=\textwidth]{CV_EnzeXu.pdf}
% \end{center}
% \clearpage
% % \includepdf[pages=-,scale=0.6]{CV_EnzeXu.pdf}

\end{document}